\documentclass{article} % For LaTeX2e
\usepackage{iclr2019_conference,times}

\usepackage{amsmath,amsfonts,bm}

\def\eqref#1{equation~\ref{#1}}
\def\1{\bm{1}}

\def\mPhi{{\bm{\Phi}}}

\DeclareMathAlphabet{\mathsfit}{\encodingdefault}{\sfdefault}{m}{sl}
\SetMathAlphabet{\mathsfit}{bold}{\encodingdefault}{\sfdefault}{bx}{n}

\def\sN{{\mathbb{N}}}
\def\sO{{\mathbb{O}}}
\def\sP{{\mathbb{P}}}

\usepackage{hyperref}
\usepackage{url}

\usepackage[utf8]{inputenc} % allow utf-8 input
\usepackage[T1]{fontenc}    % use 8-bit T1 fonts
\usepackage{hyperref}       % hyperlinks
\usepackage{url}            % simple URL typesetting
\usepackage{booktabs}       % professional-quality tables
\usepackage{amsfonts}       % blackboard math symbols
\usepackage{nicefrac}       % compact symbols for 1/2, etc.
\usepackage{microtype}      % microtypography

\usepackage{wrapfig}
\usepackage{tabularx}
\usepackage{graphicx}
\title{Learning concise representations for regression by evolving networks of trees}

\author{
    William La Cava\thanks{Send correspondence to \texttt{lacava@upenn.edu}}, 
    Tilak Raj Singh,  
    James P. Taggart, 
    Srinivas Suri, \&  
    Jason H. Moore
    \\
  Institute for Biomedical Informatics\\
  University of Pennsylvania\\
  \texttt{\{lacava, moore\}@upenn.edu}, \\
  \texttt{\{tilakraj, jtagg, surisr\}@seas.upenn.edu }
}

\iclrfinalcopy % Uncomment for camera-ready version, but NOT for submission.
\begin{document}

\maketitle

\begin{abstract}
    We propose and study a method for learning interpretable representations for the task of regression. Features are represented as networks of multi-type expression trees comprised of activation functions common in neural networks in addition to other elementary functions. Differentiable features are trained via gradient descent, and the performance of features in a linear model is used to weight the rate of change among subcomponents of each representation. The search process maintains an archive of representations with accuracy-complexity trade-offs to assist in generalization and interpretation. We compare several stochastic optimization approaches within this framework. We benchmark these variants on 100 open-source regression problems in comparison to state-of-the-art machine learning approaches. Our main finding is that this approach produces the highest average test scores across problems while producing representations that are orders of magnitude smaller than the next best performing method (gradient boosting). We also report a negative result in which attempts to directly optimize the disentanglement of the representation result in more highly correlated features.  
\end{abstract}

\section{Introduction}

The performance of a machine learning (ML) model depends primarily on the data representation used in training~\citep{bengio_representation_2013}, and for this reason the representational capacity of neural networks (NN)  is considered a central factor in their success in many applications~\citep{goodfellow_deep_2016}.  To date, there does not seem to be a consensus on how the architecture should be designed. As problems grow in complexity, the networks proposed to tackle them grow as well, leading to an intractable design space. One design approach is to tune network architectures through network hyperparameters using grid search or randomized search~\citep{bergstra_random_2012} with cross validation. Often some combination of hyperparameter tuning and manual design by expertise/intuition is done~\citep{goodfellow_deep_2016}. Many approaches to network architecture search exist, including weight sharing~\citep{pham_efficient_2018} and reinforcement learning~\citep{zoph_neural_2016}. Another potential solution explored in this work (and others) is to use population-based stochastic optimization (SO) methods, also known as metaheuristics~\citep{luke_essentials_2013}. In SO, several candidate solutions are evaluated and varied over several iterations, and heuristics are used to select and update the candidate networks until the population produces a desirable architecture. The general approach has been studied at least since the late 80s in various forms~\citep{miller_designing_1989,yao_evolving_1999,stanley_evolving_2002} for NN design, with several recent applications~\citep{real_large-scale_2017,jaderberg_population_2017,conti_improving_2017,real_using_2018}.

In practice, the adequacy of the architecture is often dependent on conflicting objectives. For example, interpretability may be a central concern, because many researchers in the scientific community rely on ML models not only to provide predictions that match data from various processes, but to provide insight into the nature of the processes themselves. Approaches to interpretability can be roughly grouped into semantic and syntactic approaches. Semantic approaches encompass methods that attempt to elucidate the behavior of a model under various input conditions as a way of explanation (e.g.~\citep{ribeiro_why_2016}). Syntactic methods instead focus on the development of concise models that offer insight by virtue of their simplicity, in a similar vein to models built from first-principles (e.g.~\citep{tibshirani_regression_1996,schmidt_distilling_2009}). Akin to the latter group, our goal is to discover the simplest description of a process whose predictions generalize as well as possible. 

Good representations should also disentangle the factors of variation~\citep{bengio_representation_2013} in the data, in order to ease model interpretation. Disentanglement implies functional modularity; i.e., sub-components of the network should encapsulate behaviors that model a sub-process of the task. In this sense, stochastic methods such as evolutionary computation (EC) appear well-motivated, as they are premised on the identification and propagation of building blocks of solutions~\citep{holland_adaptation_1975}. Experiments with EC applied to networks suggest it pressures networks to be modular~\citep{huizinga_evolving_2014,kashtan_spontaneous_2005}. Although the identification functional building blocks of solutions sounds ideal, we have no way of knowing \textit{a priori} whether a given problem will admit the identification of building blocks of solutions via heuristic search~\citep{oppacher_troubling_2014}. Our goal in this paper is thus to empirically assess the performance of several SO approaches in a system designed to produce intelligible representations from NN building blocks for regression. 

In Section 2, we introduce a new method for optimizing representations that we call the feature engineering automation tool (FEAT)\footnote{\url{http://github.com/lacava/feat}}. The purpose of this method is to optimize an archive of representations that characterize the trade-off between conciseness and accuracy among representations. Algorithmically, two aspects of the method distinguish FEAT from previous work. First, it represents the internal structure of each NN as a set of syntax trees, with the goal of improving the transparency of the resultant representations. Second, it uses weights learned via gradient descent to provide feedback to the variation process at a more granular level. We compare several multi-objective variants of this approach using EC and non-EC methods with different sets of objectives.

We discuss related work in more detail in Section 3. In section 4 and 5, we describe and conduct an experiment that benchmarks FEAT against state-of-the-art ML methods on 100 open-source regression problems. Future work based on this analysis is discussed in Section 6, and additional detailed results are provided in the Appendix.   

\section{Methods}

We are interested in the task of regression, for which the goal is to build a predictive model $\hat{y}(\mathbf{x})$ using $N$ paired examples $\mathcal{T} = \{(\mathbf{x}_i,y_i)\}_{i = 1}^{N}$. The regression model $\hat{y}(\mathbf{x})$ associates the inputs $\mathbf{x} \in \mathbb{R}^d$ with a real-valued output $y \in \mathbb{R}$ . The goal of feature engineering / representation learning is to find a new representation of $\mathbf{x}$ via a $m$-dimensional feature mapping $\mathbf{\phi}(\mathbf{x}): \mathbb{R}^d \rightarrow \mathbb{R}^m$, such that a model $\hat{y}(\mathbf{\phi}(\mathbf{x}))$ outperforms the model $\hat{y}(\mathbf{x})$. We will assume that each predictor in $\mathbf{x}$ is scaled to zero mean, unit-variance. 

When applying a NN to a traditional ML task like regression or classification, a fixed NN architecture $\mathbf{\phi}(\mathbf{x}, \theta)$, parameterized by $\theta$, is chosen and used to fit a model 

\begin{equation}
    \hat{y} = \phi(\mathbf{x}, \theta)^T \hat{\beta} \label{eq:ml}
\end{equation}

In this case $\phi = [\phi_1\;\dots\;\phi_m]^T$ is a NN representation with $m$ nodes in the final hidden layer and a linear output layer with estimated coefficients $\hat{\beta} = [\hat{\beta}_1\;\dots\;\hat{\beta}_m]^T$. Typically the problem is then cast as a parameter optimization problem that minimizes a loss function via gradient descent. In order to tune the structure of the representation, we instead wish to solve the joint optimization problem  

\begin{equation}\label{eq:so}
    \mathbf{\phi}^*(\mathbf{x},\theta^*) = \arg \min_{\mathbf{\phi} \in \mathbb{S}, \theta } \sum_{i}^N{L(y_i, \hat{y}_i, {\phi}, \theta,\hat{\beta})}
\end{equation}

where $\hat{\phi}(\mathbf{x},\hat{\theta})$ is chosen to minimize a cost function $L$, with global optimum $\phi^*(\mathbf{x},\theta^*)$. ($L$ may depend on $\theta$ and $\beta$ in the case of regularization.) $\mathbb{S}$ is the space of possible representations realizable by the search procedure, and $\mathbf{\phi}^*$ is the true structure of the process underlying the data. The assumption of SO approaches such as evolutionary computation (EC) and simulated annealing (SA) is that candidate solutions in $\mathbb{S}$ that are similar to each other, i.e. reachable in few mutations, are more likely to have similar costs than candidate solutions that are far apart. In these cases, despite Eqn.~\ref{eq:so} being non-convex, $\mathbb{S}$ can be effectively searched by maintaining and updating a population of candidate representations that perform well. Multi-objective SO methods extend Eqn.~\ref{eq:so} to minimizing additional cost functions~\citep{schoenauer_fast_2000}.

\subsection{FEAT}\label{s:feat}

FEAT uses a typical $\mu$ + $\lambda$ evolutionary updating scheme, where $\mu=\lambda=P$. The method optimizes a population of potential representations, $\sN = \{n_1\;\dots\;n_P\}$, where $n$ is an ``individual" in the population, iterating through these steps: 
\begin{enumerate}
    \item Fit a linear model $\hat{y} = \mathbf{x}^T\hat{\beta}$. Create an initial population $\sN$ consisting of this initial representation, $\mathbf{\phi} = \mathbf{x}$, along with $P-1$ randomly generated representations that sample $\mathbf{x}$ proportionally to $\hat{\beta}$. 
    \item While the stop criterion is not met: 
        \begin{enumerate}
            \item Select parents  $\sP \subseteq \sN$ using a selection algorithm. 
            \item Apply variation operators to parents to generate $P$ offspring $\sO$; $\sN = \sN \cup \sO$ 
            \item Reduce $\sN$ to $P$ individuals using a survival algorithm.  
        \end{enumerate}
    \item Select and return $n \in \sN$ with the lowest error on a hold-out validation set. 
    \end{enumerate}
    Individuals are evaluated using an initial forward pass, after which each representation is used to fit a linear model (Eqn.~\ref{eq:ml}) using ridge regression~\citep{hoerl_ridge_1970}. The weights of the differentiable features in the representation are then updated using stochastic gradient descent.  

\subsubsection{Representation}
\begin{wrapfigure}{L}{0.25\textwidth}
    \centering
    \includegraphics[width=0.25\textwidth]{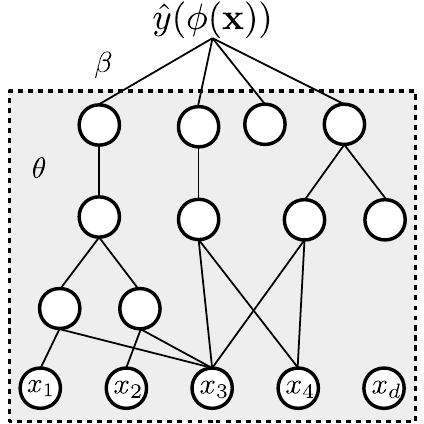}
    \caption{Example model representation in FEAT.}\label{fig:rep}
\end{wrapfigure}

The salient aspect of the proposed method is its use of syntax trees to represent the internal architecture of the network, as shown in Fig.~\ref{fig:rep}. 
FEAT constructs these trees from elementary boolean- and continuous-valued functions and literals (see Table~\ref{tbl:ops}). 
This scheme is inspired by symbolic regression (SR)~\citep{koza_genetic_1992}. 
In contrast to typical SR, each individual $n$ is a \textit{set} of such trees, the output of which is interpreted as a candidate representation, i.e. $\phi(\mathbf{x}) = [\phi_1 \dots \phi_m]$ for an individual with $m$ trees.  
The second difference from traditional SR is that the weights of differentiable nodes are encoded in the edges of the graph, rather than as independent nodes. 
We include instructions typically used as activation functions used in NN, e.g. tanh, sigmoid, logit and relu nodes, elementary arithmetic and boolean operators. 
Although a fully connected feedforward NN could be represented by this construction, representations in FEAT are biased to be thinly connected. 
Our hypothesis is that by doing so, we will improve the representation's legibility without sacrificing its capacity for modelling nonlinear relationships.

\begin{table}
    \centering
    \caption{Functions and terminals used to develop representations.}\label{tbl:ops}
    \begin{tabularx}{\textwidth}{lX} \toprule
        Continuous functions    &   \{ +, -, *, /, $^2$, $^3$, $\sqrt{}$, sin, cos, exp, log, exponent, logit, tanh, gauss, relu \} \\
        Boolean functions   &   \{ and, or, not, xor, =, <, <=, >, >= \} \\
        Terminals   &   \{$\mathbf{x}$\} \\ \bottomrule
    \end{tabularx}
\end{table}

% \subsubsection{Initialization}
% FEAT begins by fitting an ML model to the original data. For the examples in this paper, we use a linear model $\hat{y} = \mathbf{x}^T\beta$ trained using linear ridge regression. The values of $\beta$ are used to set probabilities of sampling each predictor in $\mathbf{x}$ for use in a representation, according to Eqn.~\ref{eq:ml}. This initial representation, $\mathbf{\phi} = \mathbf{x}$, is introduced into the original population, along with $P-1$ randomly generated representations. 

\subsubsection{Variation}
During variation, the representations are perturbed using a set of mutation and crossover methods. FEAT chooses among 6 variation operators that are as follows. 
\textit{ Point mutation} changes a node type to a random one with matching output type and arity.
\textit{ Insert mutation} replaces a node with a randomly generated depth 1 subtree.
\textit{ Delete mutation} removes a feature or replaces a sub-program with an input node, with equal probability.
\textit{ Insert/Delete dimension} adds or removes a new feature. 
\textit{ Sub-tree crossover} replaces a sub-tree from one parent with the sub-tree of another parent.
\textit{ Dimension crossover} swaps two features between parents.
The exact probabilities of each variation operator will affect the performance of the algorithm, and others have proposed methods for adjusting these probabilities, e.g.~\cite{igel_operator_2003}. For the purposes of our study, we use each operator with uniform probability. 

\paragraph{Feedback} The use of an ML model to assess the fitness of each representation can be used to provide information about the elements of the representation that should be changed. In particular, we assume that programs in the representation with small coefficients are the best candidates for mutation and crossover. With this in mind, let $n$ be an $m$-dimensional candidate representation with associated coefficients $\beta(n) \in \mathbb{R}^m$. Let $\tilde{\beta}_i(n) = |\beta_i|/\sum_i^m|\beta_i|$. The probability of mutation for tree $i$ in $n$ is denoted $PM_i(n)$, and defined as follows: 

\begin{eqnarray}\label{eq:mutate}
    s_i(n) &=&  \exp(1-\tilde{\beta}_{i}) / \sum_i^m{\exp(1-\tilde{\beta}_i)} \nonumber \\
    PM_i(n) &=&  f s_i(n) + (1-f) \frac{1}{m} 
\end{eqnarray}
 
The normalized coefficient magnitudes $\tilde{\beta} \in [0,1]$ are used to define softmax-normalized probabilities, $s$ in Eqn.~\ref{eq:mutate}. The smaller the coefficient, the higher the probability of mutation. The parameter $f$ is used to control the amount of feedback used to weight the probabilities; $\frac{1}{m}$ in this case represents uniform probability.  Among nodes in tree $m$, mutation occurs with uniform probability. This weighting could be extended for differentiable nodes by weighting the within-tree probabilities by the magnitude of the weights associated with each node. However we expect this would yield diminishing returns. 

\subsubsection{Selection and Survival}
The selection step selects among $P$ parents those representations that will be used to generate offspring. Following variation, the population consists of $2P$ representations of parents and offspring. The survival step is used to reduce the population back to size $P$, at which point the generation is finished. In our initial study, we empirically compared five algorithms for selection and survival: 1) $\epsilon$-lexicase selection (Lex) ~\citep{la_cava_epsilon-lexicase_2016}, 2) non-dominated sorting genetic algorithm (NSGA2)~\citep{schoenauer_fast_2000}, 3) a novel hybrid algorithm using Lex for selection and NSGA2 for survival, 4) simulated annealing~\citep{kirkpatrick_optimization_1983}, and 5) random search. These comparisons are described in Appendix Section~\ref{s:selection}. We found that the hybrid algorithm (3) performed the best; it is described below. 

Parents are selected using Lex. Lex was proposed for regression problems~\citep{la_cava_epsilon-lexicase_2016, la_cava_probabilistic_2018} as an adaption of lexicase selection~\citep{spector_assessment_2012} for continuous domains. Under $\epsilon$-lexicase selection, parents are chosen by filtering the population according to randomized orderings of training samples with the $\epsilon$ threshold defined relative to the sample loss among the selection pool. This filtering strategy scales probability of selection for an individual based on the difficulty of the training cases the individual performs well on. Lex has shown strong performance among SR methods in recent tests, motivating our interest in studying it~\citep{orzechowski_where_2018}. The survival step for Lex just preserves offspring plus the best individual in the population. 

Survival is conducted using the survival sub-routine of NSGA2, a popular strategy for multi-objective optimization~\citep{schoenauer_fast_2000}. NSGA2 applies preference for survival using Pareto dominance relations.  An individual ($n_i$) is said to \textit{dominate} another ($n_j$) if, for all objectives, $n_i$ performs at least as well as $n_j$, and for at least one objective, $n_i$ strictly outperforms $n_j$. The Pareto \textit{front} is the set of individuals in $\sN$ that are non-dominated in the population and thus represent optimal trade-offs between objectives found during search. Individuals are assigned a Pareto \textit{ranking} that specifies the number of individuals that dominate them, thereby determining their proximity to the front. 

The survival step of NSGA2 begins by sorting the population according to their Pareto front ranking and choosing the lowest ranked individuals for survival. To break rank ties, NSGA2 assigns each individual a crowding distance measure, which quantifies an individual's distance to its two adjacent neighbors in objective space. If a rank level does not completely fit in the survivor pool, individuals of that rank are sorted by highest crowding distance and added in order until $P$ individuals are chosen. 

\subsubsection{Objectives}
We consider three objectives in our study corresponding to three goals: first, to reduce model error; second, to minimize complexity of the representation; and third, to minimize the entanglement of the representation. We test the third objective using two different metrics: the correlation of the transformation matrix $\phi(\mathbf{x})$ and its condition number. These metrics are defined below. 

The first objective always corresponds to the mean squared loss function for individual $n$, and the second corresponds to the complexity of the representation. There are many ways to define complexity of an expression; one could simply look at the number of operations in a representation, or look at the behavioral complexity of the representation using a polynomial order~\citep{vladislavleva_order_2009}. The one we use, which is similar to that used by~\cite{kommenda_michael_evolving_2015}, is to assign a complexity weight to each operator (see Table~\ref{tbl:ops}), with higher weights assigned to operators considered more complex. If the weight of operator $o$ is $c_o$, then the complexity of an expression tree beginning at node $o$ is defined recursively as

\begin{equation}
    C(o) = c_o \sum_{a=1}^kC(a) \label{eq:complex}
\end{equation}

where $o$ has $k$ arguments, and $C(a)$ is the complexity of argument $a$. The complexity of a representation is then defined as the sum of the complexities of its output nodes. The goal of defining complexity in such a way is to discourage deep sub-expressions within complex nodes, which are often hard to interpret. It's important to note that the choice of operator weights is bound to be subjective, since we lack an objective notion of interpretability. For this reason, although we use Eqn.~\ref{eq:complex} to drive search, our experimental comparisons with other algorithms rely on the node counts of the final models for benchmarking interpretability of different methods.

% \paragraph{Quantifying disentanglement}
{\it Disentanglement} is a term used to describe the notion of a representation's ability to separate factors of variation in the underlying process~\citep{bengio_representation_2013}. Although a thorough review is beyond the scope of this section, there is a growing body of literature addressing disentanglement, primarily with unsupervised learning and/or image analysis
~\citep{montavon_better_2012,whitney_disentangled_2016,higgins_-vae:_2017,gonzalez-garcia_image--image_2018,hadad_two-step_2018,kumar_variational_2018}. There are various ways to quantify disentanglement. For instance, \cite{brahma_why_2016} proposed measuring disentanglement as the difference between geodesic and Euclidean distances among points on a manifold (i.e. training instances). If the latent structure is known, the information-theoretic metrics proposed by \cite{eastwood_framework_2018} may be used. 
In the case of regression, a disentangled representation ideally contains a minimal set of features, each corresponding to a separate latent factor of variation, and each orthogonal to each other. In this regard, we attempt to minimize the collinearity between features in $\phi$ as a way to promote disentanglement. We tested two measurements of collinearity (a.k.a. multicollinearity) in the derived feature space.
The first is the average squared Pearson's correlation among features of $\phi$, i.e., 
\begin{equation}
    Corr(\phi) = \frac{1}{N(N-1)} \sum_{\phi_i,\phi_j \in \phi, i \neq j}{\left( \frac{\text{cov}(\phi_i,\phi_j)}{\sigma(\phi_i)\sigma{(\phi_j)}} \right)^2} \label{eq:corr}
\end{equation}
Eqn.~\ref{eq:corr} is relatively inexpensive to compute but only captures bivariate correlations in $\phi$. As a result we also test the condition number (CN). 
Consider the $N \times m$ representation matrix $\mPhi$. 
The CN of $\mPhi$ is defined as  
\begin{equation}
    CN(\phi) = \frac{\mu_{\text{max}}(\mPhi)}{\mu_{\text{min}}(\mPhi)} \label{eq:cn}
\end{equation}
where $\mu_{\text{max}}$ and $\mu_{\text{min}}$ are the largest and smallest singular values of $\mPhi$. Unlike $Corr$, $CN$ can capture higher-order dependencies in the representation. $CN$ is also related directly to the sensitivity of $\mPhi$ to perturbations in the training data~\citep{belsley_guide_1991,cline_estimate_1979}, and thus captures a notion of network invariance explored in previous work by~\cite{goodfellow_measuring_2009}. We consider another common measure of multicollinearity, the variance inflation factor~\citep{obrien_caution_2007}, to be too expensive for our purposes. 

\section{Related Work}

% \paragraph{Neuroevolution}
The idea to evolve NN architectures is well established in literature, and is known as neuroevolution. Popular methods of neuroevolution include neuroevolution of augmenting topologies (NEAT\citep{stanley_evolving_2002} and Hyper-NEAT\citep{stanley_hypercube-based_2009}), and compositional pattern producing networks~\citep{stanley_compositional_2007} . The aforementioned approaches eschew the parameter learning step common in other NN paradigms, although others have developed integrations~\citep{fernando_convolution_2016}.  In addition, they have been developed predominantly for other task domains such as robotics and control~\citep{gomez_efficient_2006}, image classification~\citep{real_large-scale_2017,real_using_2018}, and reinforcement learning~\citep{igel_neuroevolution_2003,conti_improving_2017}. Reviews of these methods are available~\citep{floreano_neuroevolution:_2008,stanley_designing_2019}.   

Most neuroevolution strategies do not have interpretability as a core focus, and thus do not attempt to use multi-objective methods to update the networks. An exception is the work of \cite{wiegand_evolutionary_2004}, in which a template NN was optimized using a multi-objective EC method with size as an objective. In this case, the goal was to reduce computational complexity in face detection.

Neuroevolution is a part of a broader research field of neural architecture search (NAS)~\citep{zoph_neural_2016,le_using_2017,liu_progressive_2017}. NAS methods vary in approach, including for example parameter sharing~\citep{pham_efficient_2018}, sequential model-based optimization~\citep{liu_progressive_2017}, reinforcement learning~\citep{zoph_neural_2016}, and greedy heuristic strategies~\citep{cortes_adanet:_2016}. 

% \paragraph{ Symbolic Regression} 
FEAT is also related to SR approaches to feature engineering~\citep{krawiec_genetic_2002,arnaldo_multiple_2014,la_cava_general_2017,la_cava_multidimensional_2018,munoz_evolving_2018} that use EC to search for possible representations and couple with an ML model to handle the parametrization of the representations. SR methods have been successful in developing intelligible models of physical systems~\citep{schmidt_distilling_2009,la_cava_automatic_2016}. 
FEAT differs from these methods in the following ways. A key challenge in SR is understanding functional modularity within representations/programs that can be exploited for search. FEAT is designed with the insight that ML weights can be leveraged during variation to promote functional building blocks, an exploit not used in previous methods. Second, FEAT uses multiple type representations, and thus can learn continuous and rule-based features within a single representation, unlike previous methods. This is made possible using a stack-based encoding with strongly-typed operators. Finally, FEAT incorporates two elements of NN learning to improve its representational capacity: activation functions commonly used in NN and edge-based encoding of weights. Traditionally, SR operates with standard mathematical operators, and treats constants as leaves in the expression trees rather than edge weights. An exception is MRGP~\citep{arnaldo_multiple_2014}, which encodes weights at each node but updates them via Lasso instead of using gradient descent with backpropagation. SR methods have also been paired with various parameter learning strategies, including those based on backpropagation~\citep{topchy_faster_2001, kommenda_effects_2013,izzo_differentiable_2017}. It should be noted that non-stochastic methods for SR exist, such as mixed integer non-linear programming, which has been demonstrated for small search spaces~\citep{austel_globally_2017}.

% \paragraph {Dropout } FEAT also shares a motivational relationship to dropout, a popular method of NN regularization. Dropout is an approach that may improve interpretability of models by considering competing subsets of networks during training\citep{srivastava_dropout:_2014}. The authors were motivated in part by the evolutionary process of sexual recombination, which is a dominant form of genotype variation found in nature for unclear reasons. One reason the authors entertain is the ability of crossover between models to assert selective pressure for genes to be robust to different environmental contexts, since crossover may introduce large changes to neighboring genes. This pressure is also rewards genes with modular functionality since genes that are close together are more likely to be shared together and must perform a similar function in a new organism. %<F6>Experimental investigations have found that changing environmental contexts are important for the development of network modularity~\citep{kashtan_spontaneous_2005} in natural systems. 

% \paragraph {Disentanglement}

%%%%%%%%%%%%%%%%%%%%%%%%%%%%%%%%%%%%%%%%%%%%%%%%%%%%%%%%%%%%%%%%%%%%%%% PMLB
\begin{wrapfigure}{R}{0.35\textwidth}
\begin{minipage}{0.35\textwidth}
    \centering
    \includegraphics[width=\textwidth]{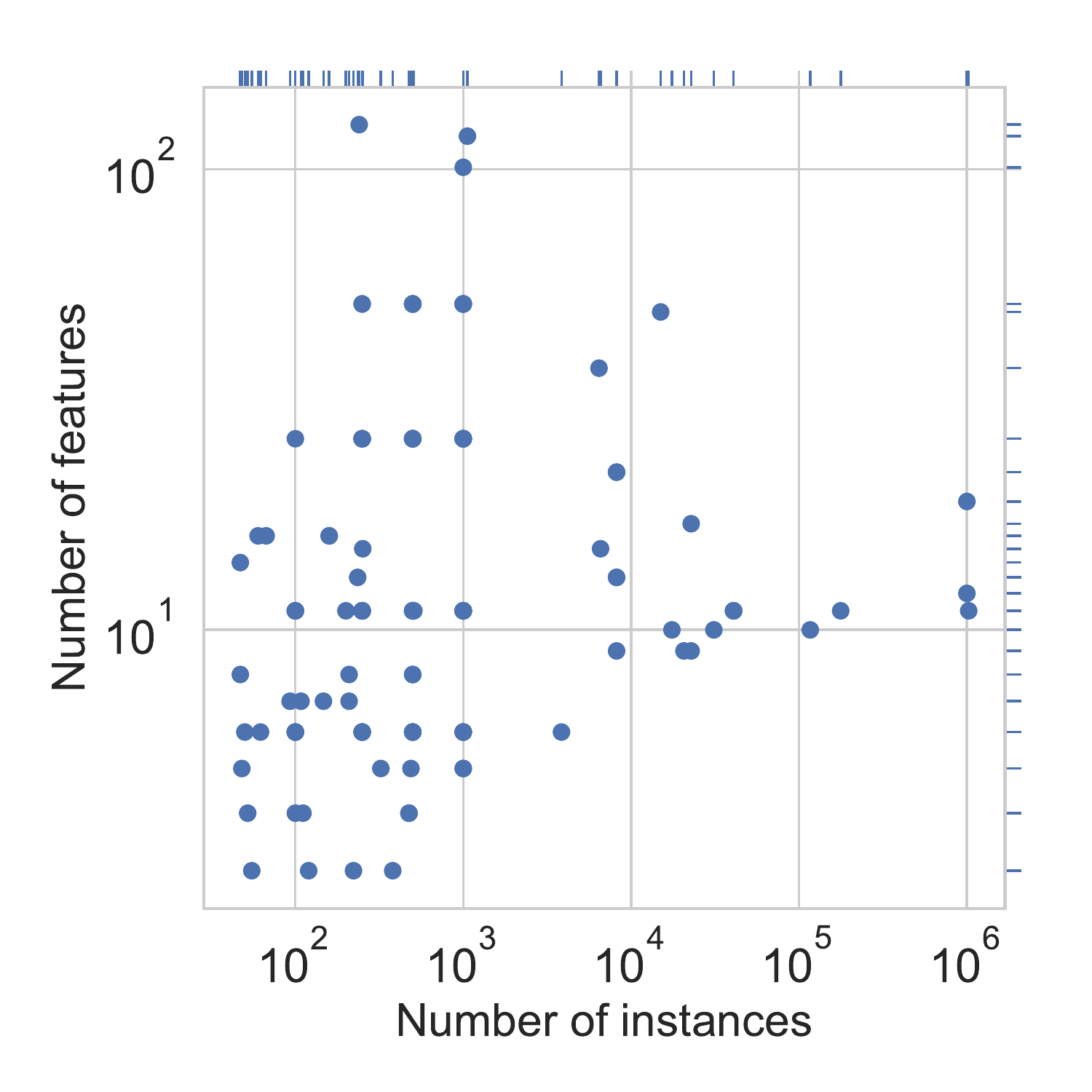}
    \caption{Properties of the regression benchmarks.}\label{fig:pmlb}
\end{minipage}
\end{wrapfigure}

\section{Experiment}\label{s:exp}
Our goals with the experiment are to 1) robustly compare FEAT to state-of-the-art regression methods, including hyperparameter optimization of feedforward NNs; 2) characterize the complexity of the models; and 3) assess whether disentanglement objectives lead to less correlated representations. For the regression datasets, we use 100 real-world and simulated datasets available from OpenML~\citep{vanschoren_openml:_2014}. The datasets are characterized in terms of number of features and sample sizes in Figure~\ref{fig:pmlb}. We use the standardized versions of the datasets available in the Penn Machine Learning Benchmark repository~\citep{olson_pmlb:_2017}. We compare the FEAT configurations to multi-layer perceptron (MLP), random forest (RF) regression, kernel ridge (KernelRidge) regression, and elastic net (ElasticNet) regression, using implementations from scikit-learn~\citep{pedregosa_scikit-learn:_2011}. In addition, we compare to XGBoost (XGB), a gradient boosting method that has performed well in recent competitions~\citep{chen_xgboost:_2016}.  Code to reproduce these experiments is available online.\footnote{\url{https://github.com/lacava/iclr_2019}} 

Each method's hyperparameters are tuned according to Table~\ref{tbl:ho} in Appendix~\ref{s:extend_exp}. For FEAT, we limit optimization to 200 iterations or 60 minutes, whichever comes first. We also stop sooner if the median validation fitness stops improving. For each method, we use grid search to tune the hyperparameters with 10-fold cross validation (CV). We use the $R^2$ CV score for assessing performance. In our results we report the CV scores for each method using its best hyperparameters. The algorithms are ranked on each dataset using their median CV score over 5 randomized shuffles of the dataset. For comparing complexity, we count the number of nodes in the final model produced by each method for each trial on each dataset. To quantify the "entanglement" of the feature spaces, we report Eqn.~\ref{eq:corr} in the raw data and in the final hidden layer of FEAT and MLP models. We also test two additional versions of Feat, denoted FeatCorr and FeatCN, that include a third objective corresponding to Eqn.~\ref{eq:corr} and~\ref{eq:cn}, respectively. %The correlation coefficient matrix represents the pairwise covariance of feature columns. We shift these between zero and 0 and compute the mean.  

Finally, we examine the FEAT results in detail for one of the benchmark datasets. For this dataset we plot the final population of models, illustrate model selection and compare the resultant features to results from linear and ensemble tree-based results. This gives practical insight into the method and provides a sense of the intelligibility of an example representation.  

\section{Results}\label{s:res}
% scores
The score statistics for each method are shown in Fig.~\ref{fig:score} across all datasets. Full statistical comparisons are reported in Appendix~\ref{s:stats}. Over all, FEAT and XGBoost produce the best predictive performance across datasets without significant differences between the two ($p$=1.0). FEAT significantly outperforms MLP, RF, KernelRidge and ElasticNet ($p \leq$1.18e-4), as does XGBoost ($p\leq$1.6e-3). 

% size
% TODO: add p-value for size
As measured by the number of nodes in the final solutions, the models produced by FEAT are significantly less complex than XGBoost, RF, and MLP, as shown in Fig.~\ref{fig:size} ($p$<1e-16). FEAT's final models tend to be within 1 order of magnitude of the linear models (ElasticNet), and 2-4 orders of magnitude smaller than the other non-linear methods.

% time
A comparison of wall-clock time is given in Fig.~\ref{fig:time} in the appendix. FEAT and MLP take approximately the same time to run, followed by XGBoost, RF, KernelRidge, and ElasticNet, in that order.    

%%%%%%%%%%%%%%%%%%%%%%%%%%%%%%%%%%%%%%%%%%%%%%%%%%%%%%%%%%%%%%%%%%%%%%% Scores
\begin{figure}
\begin{minipage}{0.49\textwidth}    
    
    \centering
    \includegraphics[width=\textwidth]{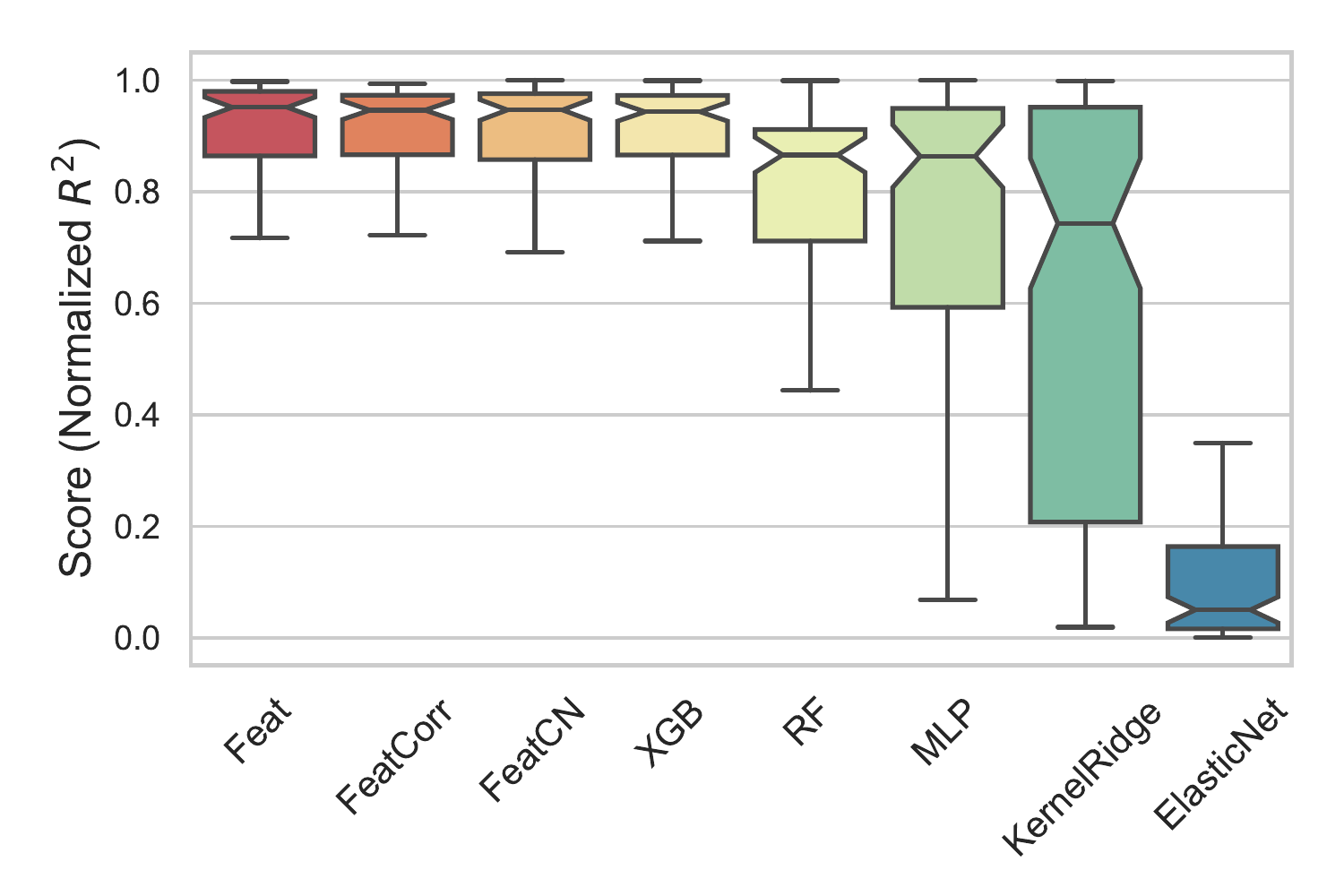}
    \caption{Mean 10-fold CV $R^2$ performance for various SO methods in comparison to other ML methods, across the benchmark problems.}\label{fig:score}

\end{minipage}
\hspace{0.01\textwidth}
%%%%%%%%%%%%%%%%%%%%%%%%%%%%%%%%%%%%%%%%%%%%%%%%%%%%%%%%%%%%%%%%%%%%%%% Size
\begin{minipage}{0.49\textwidth}
    \centering    
    \includegraphics[width=\textwidth]{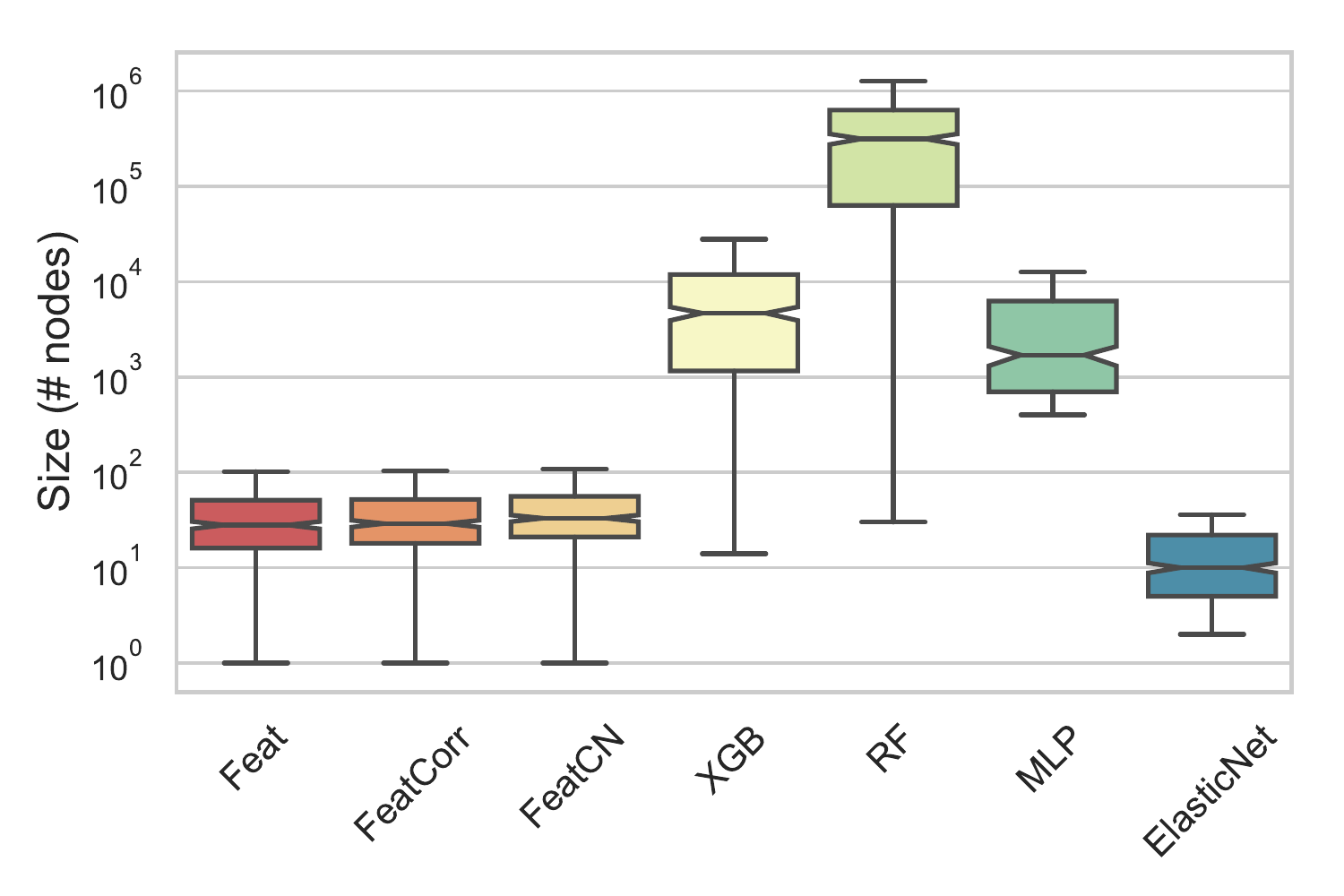}
    \caption{Size comparisons of the final models in terms of number of nodes in the solutions. }\label{fig:size}
\end{minipage}
\end{figure}
%%%%%%%%%%%%%%%%%%%%%%%%%%%%%%%%%%%%%%%%%%%%%%%%%%%%%%%%%%%%%%%%%%%%%%% Correlation
\begin{wrapfigure}{R}{0.4\textwidth}
    \centering
    \includegraphics[width=0.4\textwidth]{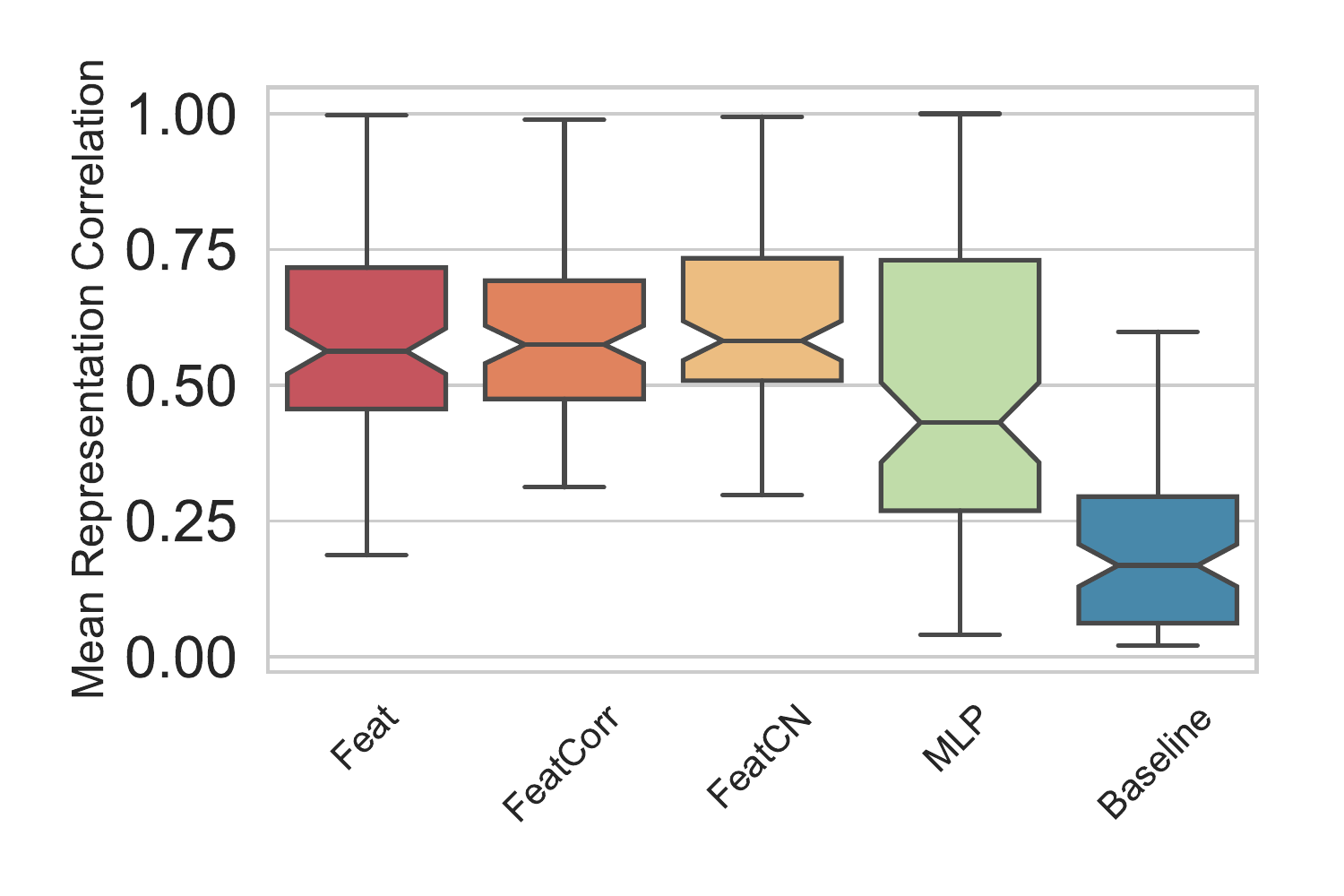}
    \caption{Mean pairwise correlations for representations produced by different methods. Baseline refers to the correlations in the original data. }\label{fig:corr}
\end{wrapfigure}
% entanglement
Fig.~\ref{fig:corr} shows the average pairwise correlations of the representations produced by Feat variants and MLP in comparison to the correlation structure of the original data. In general, MLP and FEAT tend to produce correlated feature spaces, and Feat's representations tend to contain more bivariate correlations than MLP. Furthermore, the results suggest that explicitly minimizing collinearity (FeatCorr and FeatCN) tends to produce representations that exhibit equivalent or higher levels of correlation. This result conflicts with our hypothesis, and is discussed more in Section~\ref{s:conc}.

\begin{figure}
    
    \begin{minipage}{0.65\textwidth}
    \centering
    \includegraphics[width=\textwidth]{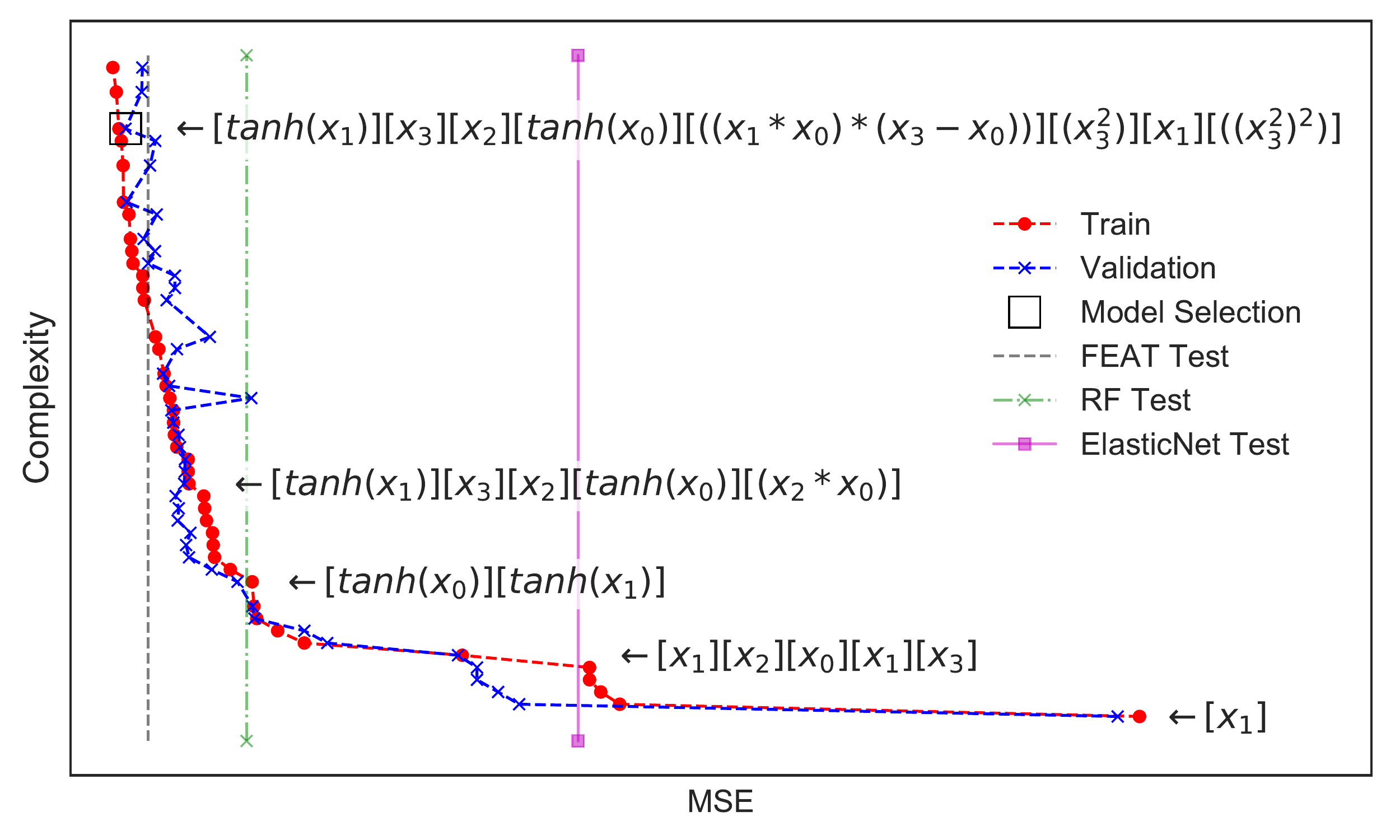}
    \end{minipage}
    % \hspace{0.01em}
    \begin{minipage}{0.35\textwidth}
        % \begin{table}
        \centering
        \begin{tabular}{l r}
                $\phi(\mathbf{x})$ & $\beta$ \\ \toprule
				$\tanh(x_1)$ & 2953.8 \\
				$x_1$ & 1961.5 \\
				$(x_3^2)$ & 329.0 \\
				$\tanh(x_0)$ & -318.5 \\
				$((x_1 \cdot x_0) \cdot (x_3-x_0))$ & 303.6 \\
				$((x_3^2)^2)$ & -288.7 \\
				$x_3$ & 203.3 \\
				$x_2$ & 64.8 \\       
				\bottomrule
		\end{tabular}
    \end{minipage}

    \caption{(Left) Representation archive for the visualizing galaxies dataset. (Right) Selected model and its weights. Internal weights omitted.}\label{fig:archive}
\end{figure}

\paragraph{Illustrative Example} We show an illustrative example of the final archive and model selection process from applying FEAT to a galaxy visualization dataset~\citep{cleveland_visualizing_1993} in Figure~\ref{fig:archive}. The red and blue points correspond to training and validation scores for each archived representation with a square denoting the final model selection. Five of the representations are printed in plain text, with each feature separated by brackets. The vertical lines in the left figure denote the test scores for FEAT, RF and ElasticNet. It is interesting to note that ElasticNet performance roughly matches the performance of a linear representation, and the RF test performance corresponds to the representation $[\tanh(x_0)][\tanh(x_1)]$ that is suggestive of axis-aligned splits for $x_0$ and $x_1$. The selected model is shown on the right, with the features sorted according to the magnitudes of $\beta$ in the linear model. The final representation combines tanh, polynomial, linear and interacting features. This representation is a clear extension of simpler ones in the archive, and the archive thereby serves to characterize the improvement in predictive accuracy brought about by increasing complexity. Although a mechanistic interpretation requires domain expertise, the final representation is certainly concise and amenable to interpretation. 

% \hspace{0.02\textwidth}
%%%%%%%%%%%%%%%%%%%%%%%%%%%%%%%%%%%%%%%%%%%%%%%%%%%%%%%%%%%%%%%%%%%%%%% Ranks
% \begin{minipage}{0.49\textwidth}
%         \centering
%     \includegraphics[width=\textwidth]{figs/barplot_ranks.pdf}
%     \caption{Rankings of each method according to median 10-fold CV $R^2$ performance, averaged over the datasets.}\label{fig:ranks}
% \end{minipage}

%%%%%%%%%%%%%%%%%%%%%%%%%%%%%%%%%%%%%%%%%%%%%%%%%%%%%%%%%%%%%%%%%%%%%%%%%%%%%%%%%%%%%%%%%%%%%%%%%%%% Archive
\section{Discussion and Conclusion}\label{s:conc}
This paper proposes a feature engineering archive tool that optimizes neural network architectures by representing them as syntax trees. FEAT uses model weights as feedback to guide network variation in an EC optimization algorithm. We conduct a thorough analysis of this method applied to the task of regression in comparison to state-of-the-art methods. The results suggest that FEAT achieves state-of-the-art performance on regression tasks while producing representations that are significantly less complex than those resulting from similarly performing methods. 
This improvement comes at an additional computational cost, limited in this study to 60 minutes per training instance. We expect this limitation to be reasonable for many applications where intelligibility is the prime motivation. 

Future work should consider the issue of representation disentanglement in more depth. Our attempts to include additional search objectives that explicitly minimize multicollinearity were not successful. Although more analysis is needed to confirm this, we suspect that the model selection procedure (Section~\ref{s:feat}, step 3) permits highly collinear representations to be chosen. This is because multicollinearity primarily affects the standard errors of $\hat{\beta}$ \citep{belsley_guide_1991}, and is not necessarily detrimental to validation error. Therefore it could be incorrect to expect the model selection procedure to effectively choose more disentangled representations. Besides improving the model selection procedure, it may be fruitful to pressure disentanglement at other stages of the search process. For example, the variation process could prune highly correlated features, or the disentanglement metric could be combined with error into a single loss function with a tunable parameter. We hope to pursue these ideas in future studies. 

\section{Acknowledgments}
This work was supported by NIH grants AI116794 and LM012601.

\bibliographystyle{iclr2019_conference}
\bibliography{iclr_2019}

\begin{thebibliography}{71}
\providecommand{\natexlab}[1]{#1}
\providecommand{\url}[1]{\texttt{#1}}
\expandafter\ifx\csname urlstyle\endcsname\relax
  \providecommand{\doi}[1]{doi: #1}\else
  \providecommand{\doi}{doi: \begingroup \urlstyle{rm}\Url}\fi

\bibitem[Arnaldo et~al.(2014)Arnaldo, Krawiec, and
  O'Reilly]{arnaldo_multiple_2014}
Ignacio Arnaldo, Krzysztof Krawiec, and Una-May O'Reilly.
\newblock Multiple regression genetic programming.
\newblock In \emph{Proceedings of the 2014 conference on {Genetic} and
  evolutionary computation}, pp.\  879--886. ACM Press, 2014.
\newblock ISBN 978-1-4503-2662-9.
\newblock \doi{10.1145/2576768.2598291}.
\newblock URL \url{http://dl.acm.org/citation.cfm?doid=2576768.2598291}.

\bibitem[Austel et~al.(2017)Austel, Dash, Gunluk, Horesh, Liberti, Nannicini,
  and Schieber]{austel_globally_2017}
Vernon Austel, Sanjeeb Dash, Oktay Gunluk, Lior Horesh, Leo Liberti, Giacomo
  Nannicini, and Baruch Schieber.
\newblock Globally {Optimal} {Symbolic} {Regression}.
\newblock \emph{arXiv:1710.10720 [stat]}, October 2017.
\newblock URL \url{http://arxiv.org/abs/1710.10720}.
\newblock arXiv: 1710.10720.

\bibitem[Belsley(1991)]{belsley_guide_1991}
David~A. Belsley.
\newblock A {Guide} to using the collinearity diagnostics.
\newblock \emph{Computer Science in Economics and Management}, 4\penalty0
  (1):\penalty0 33--50, February 1991.
\newblock ISSN 1572-9974.
\newblock \doi{10.1007/BF00426854}.
\newblock URL \url{https://doi.org/10.1007/BF00426854}.

\bibitem[Bengio et~al.(2013)Bengio, Courville, and
  Vincent]{bengio_representation_2013}
Yoshua Bengio, Aaron Courville, and Pascal Vincent.
\newblock Representation learning: {A} review and new perspectives.
\newblock \emph{IEEE transactions on pattern analysis and machine
  intelligence}, 35\penalty0 (8):\penalty0 1798--1828, 2013.
\newblock URL
  \url{http://ieeexplore.ieee.org/xpls/abs_all.jsp?arnumber=6472238}.

\bibitem[Bergstra \& Bengio(2012)Bergstra and Bengio]{bergstra_random_2012}
James Bergstra and Yoshua Bengio.
\newblock Random search for hyper-parameter optimization.
\newblock \emph{Journal of Machine Learning Research}, 13\penalty0
  (Feb):\penalty0 281--305, 2012.

\bibitem[Brahma et~al.(2016)Brahma, Wu, and She]{brahma_why_2016}
Pratik~Prabhanjan Brahma, Dapeng Wu, and Yiyuan She.
\newblock Why {Deep} {Learning} {Works}: {A} {Manifold} {Disentanglement}
  {Perspective}.
\newblock \emph{IEEE Trans. Neural Netw. Learning Syst.}, 27\penalty0
  (10):\penalty0 1997--2008, 2016.

\bibitem[Chen \& Guestrin(2016)Chen and Guestrin]{chen_xgboost:_2016}
Tianqi Chen and Carlos Guestrin.
\newblock {XGBoost}: {A} {Scalable} {Tree} {Boosting} {System}.
\newblock In \emph{Proceedings of the 22Nd {ACM} {SIGKDD} {International}
  {Conference} on {Knowledge} {Discovery} and {Data} {Mining}}, {KDD} '16, pp.\
   785--794, New York, NY, USA, 2016. ACM.
\newblock ISBN 978-1-4503-4232-2.
\newblock \doi{10.1145/2939672.2939785}.
\newblock URL \url{http://doi.acm.org/10.1145/2939672.2939785}.

\bibitem[Cleveland(1993)]{cleveland_visualizing_1993}
William~S Cleveland.
\newblock \emph{Visualizing data}.
\newblock Hobart Press, 1993.

\bibitem[Cline et~al.(1979)Cline, Moler, Stewart, and
  Wilkinson]{cline_estimate_1979}
A.~Cline, C.~Moler, G.~Stewart, and J.~Wilkinson.
\newblock An {Estimate} for the {Condition} {Number} of a {Matrix}.
\newblock \emph{SIAM Journal on Numerical Analysis}, 16\penalty0 (2):\penalty0
  368--375, April 1979.
\newblock ISSN 0036-1429.
\newblock \doi{10.1137/0716029}.
\newblock URL \url{https://epubs.siam.org/doi/abs/10.1137/0716029}.

\bibitem[Conti et~al.(2017)Conti, Madhavan, Such, Lehman, Stanley, and
  Clune]{conti_improving_2017}
Edoardo Conti, Vashisht Madhavan, Felipe~Petroski Such, Joel Lehman, Kenneth~O.
  Stanley, and Jeff Clune.
\newblock Improving {Exploration} in {Evolution} {Strategies} for {Deep}
  {Reinforcement} {Learning} via a {Population} of {Novelty}-{Seeking}
  {Agents}.
\newblock \emph{arXiv:1712.06560 [cs]}, December 2017.
\newblock URL \url{http://arxiv.org/abs/1712.06560}.
\newblock arXiv: 1712.06560.

\bibitem[Cortes et~al.(2016)Cortes, Gonzalvo, Kuznetsov, Mohri, and
  Yang]{cortes_adanet:_2016}
Corinna Cortes, Xavi Gonzalvo, Vitaly Kuznetsov, Mehryar Mohri, and Scott Yang.
\newblock Adanet: {Adaptive} structural learning of artificial neural networks.
\newblock \emph{arXiv preprint arXiv:1607.01097}, 2016.

\bibitem[Deb et~al.(2000)Deb, Agrawal, Pratap, and
  Meyarivan]{schoenauer_fast_2000}
Kalyanmoy Deb, Samir Agrawal, Amrit Pratap, and T~Meyarivan.
\newblock A {Fast} {Elitist} {Non}-dominated {Sorting} {Genetic} {Algorithm}
  for {Multi}-objective {Optimization}: {NSGA}-{II}.
\newblock In Marc Schoenauer, Kalyanmoy Deb, Günther Rudolph, Xin Yao, Evelyne
  Lutton, Juan~Julian Merelo, and Hans-Paul Schwefel (eds.), \emph{Parallel
  {Problem} {Solving} from {Nature} {PPSN} {VI}}, volume 1917, pp.\  849--858.
  Springer Berlin Heidelberg, Berlin, Heidelberg, 2000.
\newblock ISBN 978-3-540-41056-0.
\newblock URL \url{http://repository.ias.ac.in/83498/}.

\bibitem[Demšar(2006)]{demsar_statistical_2006}
Janez Demšar.
\newblock Statistical {Comparisons} of {Classifiers} over {Multiple} {Data}
  {Sets}.
\newblock \emph{Journal of Machine Learning Research}, 7\penalty0
  (Jan):\penalty0 1--30, 2006.
\newblock ISSN ISSN 1533-7928.
\newblock URL \url{http://www.jmlr.org/papers/v7/demsar06a.html}.

\bibitem[Eastwood \& Williams(2018)Eastwood and
  Williams]{eastwood_framework_2018}
Cian Eastwood and Christopher K.~I. Williams.
\newblock A {Framework} for the {Quantitative} {Evaluation} of {Disentangled}
  {Representations}.
\newblock February 2018.
\newblock URL \url{https://openreview.net/forum?id=By-7dz-AZ}.

\bibitem[Fernando et~al.(2016)Fernando, Banarse, Reynolds, Besse, Pfau,
  Jaderberg, Lanctot, and Wierstra]{fernando_convolution_2016}
Chrisantha Fernando, Dylan Banarse, Malcolm Reynolds, Frederic Besse, David
  Pfau, Max Jaderberg, Marc Lanctot, and Daan Wierstra.
\newblock Convolution by {Evolution}: {Differentiable} {Pattern} {Producing}
  {Networks}.
\newblock \emph{arXiv:1606.02580 [cs]}, June 2016.
\newblock URL \url{http://arxiv.org/abs/1606.02580}.
\newblock arXiv: 1606.02580.

\bibitem[Floreano et~al.(2008)Floreano, Dürr, and
  Mattiussi]{floreano_neuroevolution:_2008}
Dario Floreano, Peter Dürr, and Claudio Mattiussi.
\newblock Neuroevolution: from architectures to learning.
\newblock \emph{Evolutionary Intelligence}, 1\penalty0 (1):\penalty0 47--62,
  2008.
\newblock URL \url{http://link.springer.com/article/10.1007/s12065-007-0002-4}.

\bibitem[Gomez et~al.(2006)Gomez, Schmidhuber, and
  Miikkulainen]{gomez_efficient_2006}
Faustino Gomez, Jürgen Schmidhuber, and Risto Miikkulainen.
\newblock Efficient non-linear control through neuroevolution.
\newblock In \emph{{ECML}}, volume 4212, pp.\  654--662. Springer, 2006.
\newblock URL
  \url{http://link.springer.com/content/pdf/10.1007/11871842.pdf#page=676}.

\bibitem[Gonzalez-Garcia et~al.(2018)Gonzalez-Garcia, van~de Weijer, and
  Bengio]{gonzalez-garcia_image--image_2018}
Abel Gonzalez-Garcia, Joost van~de Weijer, and Yoshua Bengio.
\newblock Image-to-image translation for cross-domain disentanglement.
\newblock \emph{arXiv preprint arXiv:1805.09730}, 2018.

\bibitem[Goodfellow et~al.(2009)Goodfellow, Lee, Le, Saxe, and
  Ng]{goodfellow_measuring_2009}
Ian Goodfellow, Honglak Lee, Quoc~V. Le, Andrew Saxe, and Andrew~Y. Ng.
\newblock Measuring invariances in deep networks.
\newblock In \emph{Advances in neural information processing systems}, pp.\
  646--654, 2009.

\bibitem[Goodfellow et~al.(2016)Goodfellow, Bengio, and
  Courville]{goodfellow_deep_2016}
Ian Goodfellow, Yoshua Bengio, and Aaron Courville.
\newblock \emph{Deep {Learning}}.
\newblock MIT Press, 2016.

\bibitem[Hadad et~al.(2018)Hadad, Wolf, and Shahar]{hadad_two-step_2018}
Naama Hadad, Lior Wolf, and Moni Shahar.
\newblock A {Two}-{Step} {Disentanglement} {Method}.
\newblock In \emph{Proceedings of the {IEEE} {Conference} on {Computer}
  {Vision} and {Pattern} {Recognition}}, pp.\  772--780, 2018.

\bibitem[Higgins et~al.(2017)Higgins, Matthey, Pal, Burgess, Glorot, Botvinick,
  Mohamed, and Lerchner]{higgins_-vae:_2017}
Irina Higgins, Loic Matthey, Arka Pal, Christopher Burgess, Xavier Glorot,
  Matthew Botvinick, Shakir Mohamed, and Alexander Lerchner.
\newblock $\beta$-{VAE}: {LEARNING} {BASIC} {VISUAL} {CONCEPTS} {WITH} {A}
  {CONSTRAINED} {VARIATIONAL} {FRAMEWORK}.
\newblock pp.\ ~22, 2017.

\bibitem[Hoerl \& Kennard(1970)Hoerl and Kennard]{hoerl_ridge_1970}
Arthur~E. Hoerl and Robert~W. Kennard.
\newblock Ridge regression: {Biased} estimation for nonorthogonal problems.
\newblock \emph{Technometrics}, 12\penalty0 (1):\penalty0 55--67, 1970.

\bibitem[Holland(1975)]{holland_adaptation_1975}
John~H Holland.
\newblock Adaptation in natural and artificial systems. {An} introductory
  analysis with application to biology, control, and artificial intelligence.
\newblock \emph{Ann Arbor, MI: University of Michigan Press}, pp.\  439--444,
  1975.

\bibitem[Huizinga et~al.(2014)Huizinga, Clune, and
  Mouret]{huizinga_evolving_2014}
Joost Huizinga, Jeff Clune, and Jean-Baptiste Mouret.
\newblock Evolving neural networks that are both modular and regular:
  {HyperNEAT} plus the connection cost technique.
\newblock pp.\  697--704. ACM Press, 2014.
\newblock ISBN 978-1-4503-2662-9.
\newblock \doi{10.1145/2576768.2598232}.
\newblock URL \url{http://dl.acm.org/citation.cfm?doid=2576768.2598232}.

\bibitem[Igel(2003)]{igel_neuroevolution_2003}
Christian Igel.
\newblock Neuroevolution for reinforcement learning using evolution strategies.
\newblock In \emph{Evolutionary {Computation}, 2003. {CEC}'03. {The} 2003
  {Congress} on}, volume~4, pp.\  2588--2595. IEEE, 2003.
\newblock URL \url{http://ieeexplore.ieee.org/abstract/document/1299414/}.

\bibitem[Igel \& Kreutz(2003)Igel and Kreutz]{igel_operator_2003}
Christian Igel and Martin Kreutz.
\newblock Operator adaptation in evolutionary computation and its application
  to structure optimization of neural networks.
\newblock \emph{Neurocomputing}, 55\penalty0 (1-2):\penalty0 347--361, 2003.

\bibitem[Izzo et~al.(2017)Izzo, Biscani, and Mereta]{izzo_differentiable_2017}
Dario Izzo, Francesco Biscani, and Alessio Mereta.
\newblock Differentiable {Genetic} {Programming}.
\newblock In \emph{European {Conference} on {Genetic} {Programming}}, pp.\
  35--51. Springer, 2017.

\bibitem[Jaderberg et~al.(2017)Jaderberg, Dalibard, Osindero, Czarnecki,
  Donahue, Razavi, Vinyals, Green, Dunning, Simonyan, Fernando, and
  Kavukcuoglu]{jaderberg_population_2017}
Max Jaderberg, Valentin Dalibard, Simon Osindero, Wojciech~M. Czarnecki, Jeff
  Donahue, Ali Razavi, Oriol Vinyals, Tim Green, Iain Dunning, Karen Simonyan,
  Chrisantha Fernando, and Koray Kavukcuoglu.
\newblock Population {Based} {Training} of {Neural} {Networks}.
\newblock \emph{arXiv:1711.09846 [cs]}, November 2017.
\newblock URL \url{http://arxiv.org/abs/1711.09846}.
\newblock arXiv: 1711.09846.

\bibitem[Kashtan \& Alon(2005)Kashtan and Alon]{kashtan_spontaneous_2005}
Nadav Kashtan and Uri Alon.
\newblock Spontaneous evolution of modularity and network motifs.
\newblock \emph{Proceedings of the National Academy of Sciences}, 102\penalty0
  (39):\penalty0 13773--13778, September 2005.
\newblock ISSN 0027-8424, 1091-6490.
\newblock \doi{10.1073/pnas.0503610102}.
\newblock URL \url{http://www.pnas.org/content/102/39/13773}.

\bibitem[Kingma \& Ba(2014)Kingma and Ba]{kingma_adam:_2014}
Diederik~P. Kingma and Jimmy Ba.
\newblock Adam: {A} {Method} for {Stochastic} {Optimization}.
\newblock \emph{arXiv:1412.6980 [cs]}, December 2014.
\newblock URL \url{http://arxiv.org/abs/1412.6980}.
\newblock arXiv: 1412.6980.

\bibitem[Kirkpatrick et~al.(1983)Kirkpatrick, Gelatt, and
  Vecchi]{kirkpatrick_optimization_1983}
Scott Kirkpatrick, C~Daniel Gelatt, and Mario~P Vecchi.
\newblock Optimization by simulated annealing.
\newblock \emph{science}, 220\penalty0 (4598):\penalty0 671--680, 1983.

\bibitem[Kommenda et~al.(2013)Kommenda, Kronberger, Winkler, Affenzeller, and
  Wagner]{kommenda_effects_2013}
Michael Kommenda, Gabriel Kronberger, Stephan Winkler, Michael Affenzeller, and
  Stefan Wagner.
\newblock Effects of constant optimization by nonlinear least squares
  minimization in symbolic regression.
\newblock In Christian Blum, Enrique Alba, Thomas Bartz-Beielstein, Daniele
  Loiacono, Francisco Luna, Joern Mehnen, Gabriela Ochoa, Mike Preuss, Emilia
  Tantar, and Leonardo Vanneschi (eds.), \emph{{GECCO} '13 {Companion}:
  {Proceeding} of the fifteenth annual conference companion on {Genetic} and
  evolutionary computation conference companion}, pp.\  1121--1128, Amsterdam,
  The Netherlands, 2013. ACM.
\newblock \doi{doi:10.1145/2464576.2482691}.

\bibitem[Kommenda et~al.(2015)Kommenda, Kronberger, Affenzeller, Winkler, and
  Burlacu]{kommenda_michael_evolving_2015}
Michael Kommenda, Gabriel Kronberger, Michael Affenzeller, Stephan~M. Winkler,
  and Bogdan Burlacu.
\newblock Evolving {Simple} {Symbolic} {Regression} {Models} by
  {Multi}-objective {Genetic} {Programming}.
\newblock In \emph{Genetic {Programming} {Theory} and {Practice}}, volume XIV
  of \emph{Genetic and {Evolutionary} {Computation}}. Springer, Ann Arbor, MI,
  2015.

\bibitem[Koza(1992)]{koza_genetic_1992}
John~R. Koza.
\newblock \emph{Genetic {Programming}: {On} the {Programming} of {Computers} by
  {Means} of {Natural} {Selection}}.
\newblock MIT Press, Cambridge, MA, USA, 1992.
\newblock ISBN 0-262-11170-5.

\bibitem[Krawiec(2002)]{krawiec_genetic_2002}
Krzysztof Krawiec.
\newblock Genetic programming-based construction of features for machine
  learning and knowledge discovery tasks.
\newblock \emph{Genetic Programming and Evolvable Machines}, 3\penalty0
  (4):\penalty0 329--343, 2002.
\newblock URL \url{http://link.springer.com/article/10.1023/A:1020984725014}.

\bibitem[Kumar et~al.(2018)Kumar, Sattigeri, and
  Balakrishnan]{kumar_variational_2018}
Abhishek Kumar, Prasanna Sattigeri, and Avinash Balakrishnan.
\newblock Variational {Inference} of {Disentangled} {Latent} {Concepts} from
  {Unlabeled} {Observations}.
\newblock February 2018.
\newblock URL \url{https://openreview.net/forum?id=H1kG7GZAW}.

\bibitem[La~Cava \& Moore(2017)La~Cava and Moore]{la_cava_general_2017}
William La~Cava and Jason Moore.
\newblock A {General} {Feature} {Engineering} {Wrapper} for {Machine}
  {Learning} {Using} {\textbackslash}epsilon -{Lexicase} {Survival}.
\newblock In \emph{Genetic {Programming}}, pp.\  80--95. Springer, Cham, April
  2017.
\newblock \doi{10.1007/978-3-319-55696-3_6}.
\newblock URL
  \url{https://link.springer.com/chapter/10.1007/978-3-319-55696-3_6}.

\bibitem[La~Cava et~al.(2016{\natexlab{a}})La~Cava, Danai, Spector, Fleming,
  Wright, and Lackner]{la_cava_automatic_2016}
William La~Cava, Kourosh Danai, Lee Spector, Paul Fleming, Alan Wright, and
  Matthew Lackner.
\newblock Automatic identification of wind turbine models using evolutionary
  multiobjective optimization.
\newblock \emph{Renewable Energy}, 87, Part 2:\penalty0 892--902, March
  2016{\natexlab{a}}.
\newblock ISSN 0960-1481.
\newblock \doi{10.1016/j.renene.2015.09.068}.
\newblock URL
  \url{http://www.sciencedirect.com/science/article/pii/S0960148115303475}.

\bibitem[La~Cava et~al.(2016{\natexlab{b}})La~Cava, Spector, and
  Danai]{la_cava_epsilon-lexicase_2016}
William La~Cava, Lee Spector, and Kourosh Danai.
\newblock Epsilon-{Lexicase} {Selection} for {Regression}.
\newblock In \emph{Proceedings of the {Genetic} and {Evolutionary}
  {Computation} {Conference} 2016}, {GECCO} '16, pp.\  741--748, New York, NY,
  USA, 2016{\natexlab{b}}. ACM.
\newblock ISBN 978-1-4503-4206-3.
\newblock \doi{10.1145/2908812.2908898}.
\newblock URL \url{http://doi.acm.org/10.1145/2908812.2908898}.

\bibitem[La~Cava et~al.(2018{\natexlab{a}})La~Cava, Helmuth, Spector, and
  Moore]{la_cava_probabilistic_2018}
William La~Cava, Thomas Helmuth, Lee Spector, and Jason~H. Moore.
\newblock A probabilistic and multi-objective analysis of lexicase selection
  and epsilon-lexicase selection.
\newblock \emph{Evolutionary Computation}, pp.\  1--28, May 2018{\natexlab{a}}.
\newblock ISSN 1063-6560.
\newblock \doi{10.1162/evco_a_00224}.
\newblock URL \url{https://doi.org/10.1162/evco_a_00224}.

\bibitem[La~Cava et~al.(2018{\natexlab{b}})La~Cava, Silva, Danai, Spector,
  Vanneschi, and Moore]{la_cava_multidimensional_2018}
William La~Cava, Sara Silva, Kourosh Danai, Lee Spector, Leonardo Vanneschi,
  and Jason~H. Moore.
\newblock Multidimensional genetic programming for multiclass classification.
\newblock \emph{Swarm and Evolutionary Computation}, April 2018{\natexlab{b}}.
\newblock ISSN 2210-6502.
\newblock \doi{10.1016/j.swevo.2018.03.015}.
\newblock URL
  \url{http://www.sciencedirect.com/science/article/pii/S2210650217309136}.

\bibitem[Le \& Zoph(2017)Le and Zoph]{le_using_2017}
Quoc Le and Barret Zoph.
\newblock Using {Machine} {Learning} to {Explore} {Neural} {Network}
  {Architecture}, May 2017.
\newblock URL
  \url{https://ai.googleblog.com/2017/05/using-machine-learning-to-explore.html}.

\bibitem[Liu et~al.(2017)Liu, Zoph, Shlens, Hua, Li, Fei-Fei, Yuille, Huang,
  and Murphy]{liu_progressive_2017}
Chenxi Liu, Barret Zoph, Jonathon Shlens, Wei Hua, Li-Jia Li, Li~Fei-Fei, Alan
  Yuille, Jonathan Huang, and Kevin Murphy.
\newblock Progressive neural architecture search.
\newblock \emph{arXiv preprint arXiv:1712.00559}, 2017.

\bibitem[Luke(2013)]{luke_essentials_2013}
Sean Luke.
\newblock \emph{Essentials of {Metaheuristics}}.
\newblock 2nd edition, 2013.
\newblock ISBN 978-1-300-54962-8.

\bibitem[Miller et~al.(1989)Miller, Todd, and Hegde]{miller_designing_1989}
Geoffrey~F. Miller, Peter~M. Todd, and Shailesh~U. Hegde.
\newblock Designing {Neural} {Networks} using {Genetic} {Algorithms}.
\newblock In \emph{{ICGA}}, volume~89, pp.\  379--384, 1989.

\bibitem[Montavon \& Müller(2012)Montavon and Müller]{montavon_better_2012}
Grégoire Montavon and Klaus-Robert Müller.
\newblock Better {Representations}: {Invariant}, {Disentangled} and {Reusable}.
\newblock In \emph{Neural {Networks}: {Tricks} of the {Trade}}, Lecture {Notes}
  in {Computer} {Science}, pp.\  559--560. Springer, Berlin, Heidelberg, 2012.
\newblock ISBN 978-3-642-35288-1 978-3-642-35289-8.
\newblock \doi{10.1007/978-3-642-35289-8_29}.
\newblock URL
  \url{https://link.springer.com/chapter/10.1007/978-3-642-35289-8_29}.

\bibitem[Muñoz et~al.(2018)Muñoz, Trujillo, Silva, Castelli, and
  Vanneschi]{munoz_evolving_2018}
Luis Muñoz, Leonardo Trujillo, Sara Silva, Mauro Castelli, and Leonardo
  Vanneschi.
\newblock Evolving multidimensional transformations for symbolic regression
  with {M}3gp.
\newblock \emph{Memetic Computing}, September 2018.
\newblock ISSN 1865-9292.
\newblock \doi{10.1007/s12293-018-0274-5}.
\newblock URL \url{https://doi.org/10.1007/s12293-018-0274-5}.

\bibitem[Olson et~al.(2017)Olson, La~Cava, Orzechowski, Urbanowicz, and
  Moore]{olson_pmlb:_2017}
Randal~S. Olson, William La~Cava, Patryk Orzechowski, Ryan~J. Urbanowicz, and
  Jason~H. Moore.
\newblock {PMLB}: {A} {Large} {Benchmark} {Suite} for {Machine} {Learning}
  {Evaluation} and {Comparison}.
\newblock \emph{BioData Mining}, 2017.
\newblock URL \url{https://arxiv.org/abs/1703.00512}.
\newblock arXiv preprint arXiv:1703.00512.

\bibitem[Oppacher(2014)]{oppacher_troubling_2014}
Una-May O’Reilly~Franz Oppacher.
\newblock The troubling aspects of a building block hypothesis for genetic
  programming.
\newblock \emph{Foundations of Genetic Algorithms 1995 (FOGA 3)}, 3:\penalty0
  73, 2014.

\bibitem[Orzechowski et~al.(2018)Orzechowski, La~Cava, and
  Moore]{orzechowski_where_2018}
Patryk Orzechowski, William La~Cava, and Jason~H. Moore.
\newblock Where are we now? {A} large benchmark study of recent symbolic
  regression methods.
\newblock \emph{arXiv:1804.09331 [cs]}, April 2018.
\newblock \doi{10.1145/3205455.3205539}.
\newblock URL \url{http://arxiv.org/abs/1804.09331}.
\newblock arXiv: 1804.09331.

\bibitem[O’brien(2007)]{obrien_caution_2007}
Robert~M. O’brien.
\newblock A {Caution} {Regarding} {Rules} of {Thumb} for {Variance} {Inflation}
  {Factors}.
\newblock \emph{Quality \& Quantity}, 41\penalty0 (5):\penalty0 673--690,
  October 2007.
\newblock ISSN 1573-7845.
\newblock \doi{10.1007/s11135-006-9018-6}.
\newblock URL \url{https://doi.org/10.1007/s11135-006-9018-6}.

\bibitem[Pedregosa et~al.(2011)Pedregosa, Varoquaux, Gramfort, Michel, Thirion,
  Grisel, Blondel, Prettenhofer, Weiss, Dubourg, and
  {others}]{pedregosa_scikit-learn:_2011}
Fabian Pedregosa, Gaël Varoquaux, Alexandre Gramfort, Vincent Michel, Bertrand
  Thirion, Olivier Grisel, Mathieu Blondel, Peter Prettenhofer, Ron Weiss,
  Vincent Dubourg, and {others}.
\newblock Scikit-learn: {Machine} learning in {Python}.
\newblock \emph{Journal of Machine Learning Research}, 12\penalty0
  (Oct):\penalty0 2825--2830, 2011.
\newblock URL \url{http://www.jmlr.org/papers/v12/pedregosa11a.html}.

\bibitem[Pham et~al.(2018)Pham, Guan, Zoph, Le, and Dean]{pham_efficient_2018}
Hieu Pham, Melody~Y. Guan, Barret Zoph, Quoc~V. Le, and Jeff Dean.
\newblock Efficient {Neural} {Architecture} {Search} via {Parameter} {Sharing}.
\newblock \emph{arXiv preprint arXiv:1802.03268}, 2018.

\bibitem[Real(2018)]{real_using_2018}
Esteban Real.
\newblock Using {Evolutionary} {AutoML} to {Discover} {Neural} {Network}
  {Architectures}, March 2018.
\newblock URL
  \url{https://ai.googleblog.com/2018/03/using-evolutionary-automl-to-discover.html}.

\bibitem[Real et~al.(2017)Real, Moore, Selle, Saxena, Suematsu, Tan, Le, and
  Kurakin]{real_large-scale_2017}
Esteban Real, Sherry Moore, Andrew Selle, Saurabh Saxena, Yutaka~Leon Suematsu,
  Jie Tan, Quoc Le, and Alex Kurakin.
\newblock Large-{Scale} {Evolution} of {Image} {Classifiers}.
\newblock \emph{arXiv:1703.01041 [cs]}, March 2017.
\newblock URL \url{http://arxiv.org/abs/1703.01041}.
\newblock arXiv: 1703.01041.

\bibitem[Ribeiro et~al.(2016)Ribeiro, Singh, and Guestrin]{ribeiro_why_2016}
Marco~Tulio Ribeiro, Sameer Singh, and Carlos Guestrin.
\newblock Why should i trust you?: {Explaining} the predictions of any
  classifier.
\newblock In \emph{Proceedings of the 22nd {ACM} {SIGKDD} {International}
  {Conference} on {Knowledge} {Discovery} and {Data} {Mining}}, pp.\
  1135--1144. ACM, 2016.

\bibitem[Schmidt \& Lipson(2009)Schmidt and Lipson]{schmidt_distilling_2009}
Michael Schmidt and Hod Lipson.
\newblock Distilling free-form natural laws from experimental data.
\newblock \emph{Science}, 324\penalty0 (5923):\penalty0 81--85, 2009.
\newblock URL \url{http://www.sciencemag.org/content/324/5923/81.short}.

\bibitem[Spector(2012)]{spector_assessment_2012}
Lee Spector.
\newblock Assessment of problem modality by differential performance of
  lexicase selection in genetic programming: a preliminary report.
\newblock In \emph{Proceedings of the fourteenth international conference on
  {Genetic} and evolutionary computation conference companion}, pp.\  401--408,
  2012.
\newblock URL \url{http://dl.acm.org/citation.cfm?id=2330846}.

\bibitem[Stanley(2007)]{stanley_compositional_2007}
Kenneth~O. Stanley.
\newblock Compositional pattern producing networks: {A} novel abstraction of
  development.
\newblock \emph{Genetic programming and evolvable machines}, 8\penalty0
  (2):\penalty0 131--162, 2007.
\newblock URL \url{http://link.springer.com/article/10.1007/s10710-007-9028-8}.

\bibitem[Stanley \& Miikkulainen(2002)Stanley and
  Miikkulainen]{stanley_evolving_2002}
Kenneth~O. Stanley and Risto Miikkulainen.
\newblock Evolving neural networks through augmenting topologies.
\newblock \emph{Evolutionary computation}, 10\penalty0 (2):\penalty0 99--127,
  2002.
\newblock URL
  \url{http://www.mitpressjournals.org/doi/abs/10.1162/106365602320169811}.

\bibitem[Stanley et~al.(2009)Stanley, D'Ambrosio, and
  Gauci]{stanley_hypercube-based_2009}
Kenneth~O. Stanley, David~B. D'Ambrosio, and Jason Gauci.
\newblock A hypercube-based encoding for evolving large-scale neural networks.
\newblock \emph{Artificial life}, 15\penalty0 (2):\penalty0 185--212, 2009.
\newblock URL
  \url{http://www.mitpressjournals.org/doi/abs/10.1162/artl.2009.15.2.15202}.

\bibitem[Stanley et~al.(2019)Stanley, Clune, Lehman, and
  Miikkulainen]{stanley_designing_2019}
Kenneth~O. Stanley, Jeff Clune, Joel Lehman, and Risto Miikkulainen.
\newblock Designing neural networks through neuroevolution.
\newblock 1\penalty0 (1):\penalty0 24, 2019.
\newblock ISSN 2522-5839.
\newblock \doi{10.1038/s42256-018-0006-z}.
\newblock URL \url{https://www.nature.com/articles/s42256-018-0006-z}.

\bibitem[Tibshirani(1996)]{tibshirani_regression_1996}
Robert Tibshirani.
\newblock Regression shrinkage and selection via the lasso.
\newblock \emph{Journal of the Royal Statistical Society. Series B
  (Methodological)}, pp.\  267--288, 1996.
\newblock URL \url{http://www.jstor.org/stable/2346178}.

\bibitem[Topchy \& Punch(2001)Topchy and Punch]{topchy_faster_2001}
Alexander Topchy and William~F. Punch.
\newblock Faster genetic programming based on local gradient search of numeric
  leaf values.
\newblock In \emph{Proceedings of the {Genetic} and {Evolutionary}
  {Computation} {Conference} ({GECCO}-2001)}, pp.\  155--162, 2001.
\newblock URL \url{http://garage.cse.msu.edu/papers/GARAGe01-07-01.pdf}.

\bibitem[Vanschoren et~al.(2014)Vanschoren, van Rijn, Bischl, and
  Torgo]{vanschoren_openml:_2014}
Joaquin Vanschoren, Jan~N. van Rijn, Bernd Bischl, and Luis Torgo.
\newblock {OpenML}: {Networked} {Science} in {Machine} {Learning}.
\newblock \emph{SIGKDD Explor. Newsl.}, 15\penalty0 (2):\penalty0 49--60, June
  2014.
\newblock ISSN 1931-0145.
\newblock \doi{10.1145/2641190.2641198}.
\newblock URL \url{http://doi.acm.org/10.1145/2641190.2641198}.

\bibitem[Vladislavleva et~al.(2009)Vladislavleva, Smits, and den
  Hertog]{vladislavleva_order_2009}
E.J. Vladislavleva, G.F. Smits, and D.~den Hertog.
\newblock Order of {Nonlinearity} as a {Complexity} {Measure} for {Models}
  {Generated} by {Symbolic} {Regression} via {Pareto} {Genetic} {Programming}.
\newblock \emph{IEEE Transactions on Evolutionary Computation}, 13\penalty0
  (2):\penalty0 333--349, 2009.
\newblock ISSN 1089-778X.
\newblock \doi{10.1109/TEVC.2008.926486}.

\bibitem[Whitney(2016)]{whitney_disentangled_2016}
William Whitney.
\newblock Disentangled {Representations} in {Neural} {Models}.
\newblock \emph{arXiv:1602.02383 [cs]}, February 2016.
\newblock URL \url{http://arxiv.org/abs/1602.02383}.
\newblock arXiv: 1602.02383.

\bibitem[Wiegand et~al.(2004)Wiegand, Igel, and
  Handmann]{wiegand_evolutionary_2004}
Stefan Wiegand, Christian Igel, and Uwe Handmann.
\newblock Evolutionary multi-objective optimisation of neural networks for face
  detection.
\newblock \emph{International Journal of Computational Intelligence and
  Applications}, 4\penalty0 (03):\penalty0 237--253, 2004.

\bibitem[Yao(1999)]{yao_evolving_1999}
Xin Yao.
\newblock Evolving artificial neural networks.
\newblock \emph{Proceedings of the IEEE}, 87\penalty0 (9):\penalty0 1423--1447,
  1999.

\bibitem[Zoph \& Le(2016)Zoph and Le]{zoph_neural_2016}
Barret Zoph and Quoc~V. Le.
\newblock Neural {Architecture} {Search} with {Reinforcement} {Learning}.
\newblock November 2016.
\newblock URL \url{https://arxiv.org/abs/1611.01578}.

\end{thebibliography}

%%%%%%%%%%%%%%%%%%%%%%%%%%%%%%%%%%%%%%%%%%%%%%%%%%%%%%%%%%%%%%%%%%%%%%%%%%%%%%%%%%%%%%%%%%%%%%%%%%%%%
%                                                 APPENDIX
%%%%%%%%%%%%%%%%%%%%%%%%%%%%%%%%%%%%%%%%%%%%%%%%%%%%%%%%%%%%%%%%%%%%%%%%%%%%%%%%%%%%%%%%%%%%%%%%%%%%%
\appendix 
\section{Appendix}
\subsection{Additional Experiment Information}\label{s:extend_exp}
Table~\ref{tbl:ho} details the hyperparameters for each method used in the experimental results described in Sections~\ref{s:exp} and~\ref{s:res}.  

\begin{table}[htb!]
\footnotesize
\centering
\caption{\small Comparison methods and their hyperparameters. Tuned values denoted with brackets.}\label{tbl:ho}
\begin{tabularx}{\textwidth}{p{6em} r X}
    Method  & Setting	                    &	Value                                \\
\toprule
FEAT		& 	Population size                             &	500	\\
			& 	Termination criterion                       &	200 generations, 60 minutes, or 50 iterations of stalled median validation loss	\\
			& 	Max depth                                   &	10	\\
			& 	Max dimensionality                          &	50	\\
			& 	Objectives                                  &	\{(MSE,$C$), (MSE,$C$,$Corr$),(MSE,$C$,$CN$) \}\\
            &   Feedback ($f$)                              &   \{ 0.25, 0.5, 0.75 \} \\
            &   Crossover/mutation ratio                    &   \{ 0.25, 0.5, 0.75 \}   \\
            &   Batch size                                  & 1000                  \\
            & 	Learning rate (initial)                     & 0.1 \\
			& 	SGD iterations / individual / generation    & 10 \\
\midrule
MLP         &    Optimizer                   &   \{LBFGS, Adam~\citep{kingma_adam:_2014}\}			\\ 
            &    Hidden Layers               &	\{1,3,6\}                                   	\\
            &    Neurons                     &	\{(100,), (100,50,10), (100,50,20,10,10,8)\}	\\
            &    Learning rate               &   (initial) \{1e-4, 1e-3, 1e-2\}                  \\
            &    Activation                  &   \{logistic, tanh, relu\}                      	\\
            &    Regularization              &   $L_2$, $\alpha$ =  \{1e-5, 1e-4, 1e-3\}      	\\
            &    Max Iterations              &   10000                                       	\\
            &    Early Stopping              &   True                                        	\\
\midrule
XGBoost    
            &   Number of estimators        &   \{10, 100, 200, 500, 1000\}                          \\
            &   Max depth                   &   \{3, 4, 5, 6, 7\}                                         \\    
            &   Min split loss ($\gamma$)      &   \{1e-3,1e-2,0.1,1,10,1e2,1e3\}                  \\
            &   Learning rate               &   \{0, 0.01, \dots, 1.0 \}                        \\
\midrule
Random Forest
            &   Number of estimators        &   \{10, 100, 1000\}                                \\
            &   Min weight fraction leaf    &   \{ 0.0, 0.25, 0.5 \}                            \\
\midrule
Kernel Ridge
            &   Kernel                      &   Radial basis function                           \\
            &   Regularization ($\alpha$)   &   \{ 1e-3, 1e-2, 0.1, 1 \}                        \\
            &   Kernel width ($\gamma$)     &   \{ 1e-2, 0.1, 1, 10, 100 \}                     \\
\midrule
ElasticNet
            &   $l_1$-$l_2$ ratio       &   \{ 0, 0.01, \dots, 1.0 \}                       \\
            &   selection                   &   \{ cyclic, random \}                            \\ 
\bottomrule
\end{tabularx}
\end{table}

\begin{figure}[htb!]
    \centering
    \includegraphics[width=0.5\textwidth]{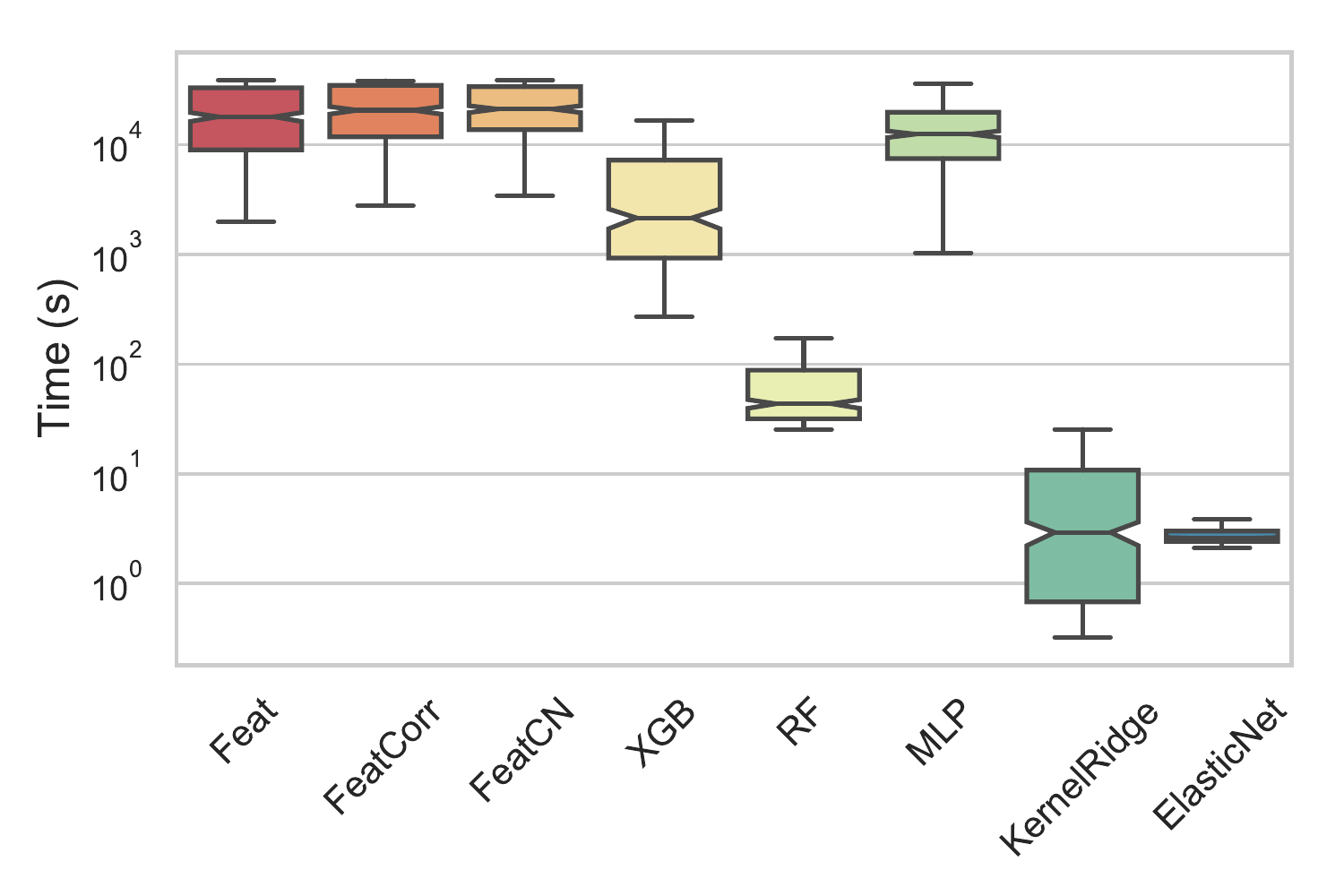}
    \caption{Wall-clock runtime for each method in seconds. }\label{fig:time}
\end{figure}

% \subsection{Wall-clock runtime}
Runs are conducted in a heterogenous computing environment, with one core assigned to each CV training per dataset. As such, wall-clock times are a flawed measure of computational complexity. With this caveat in mind, we report the wall-clock run times for each method in Fig.~\ref{fig:time}. The Feat variants are terminated at 200 generations or 60 minutes, which explains their uniformity.  Note that methods are unable to take advantage of parallelization in this experiment. 

%%%%%%%%%%%%%%%%%%%%%%%%%%%%%%%%%%%%%%%%%%%%%%%%%%%%%%%%%%%%%%%%%%%%%%% Time

\subsection{Comparison of stochastic optimization approaches}\label{s:selection}
Our initial analysis sought to determine how different SO approaches performed within this framework. We tested five methods: 1) NSGA2, 2) Lex, 3) LexNSGA2, 4) Simulated annealing, and 5) random search. The simulated annealing and random search approaches are described below. 

\paragraph {Simulated annealing} Simulated annealing (SimAnn) is a non-evolutionary technique that instead models the optimization process on the metallurgical process of annealing. In our implementation, offspring compete with their parents; in the case of multiple parents, offspring compete with the program with which they share more nodes. The probability of an offspring replacing its parent in the population is given by the equation

\begin{equation}
    P_{sel}(n_o | n_p, t) = \exp{\left(\frac{F(n_p) - F(n_o)}{t}\right)}\label{eq:SA}
\end{equation}

The probability of offspring replacing its parent is a function of its fitness, $F$, in our case the mean squared loss of the candidate model. In Eqn.~\ref{eq:SA}, $t$ is a scheduling parameter that controls the rate of ``cooling", i.e. the rate at which steps in the search space that are worse are tolerated by the update rule. In accordance with~\citep{kirkpatrick_optimization_1983}, we use an exponential schedule for $t$, defined as 
$t_{g} = (0.9)^gt_0$
, where $g$ is the current generation and $t0$ is the starting temperature. $t0$ is set to 10 in our experiments. 

\paragraph {Random search} We compare the selection and survival methods to random search, in which no assumptions are made about the structure of the search space. To conduct random search, we randomly sample $\mathbb{S}$ using the initialization procedure. Since FEAT begins with a linear model of the process, random search will produce a representation at least as good as this initial model on the internal validation set. 

\paragraph{A note on archiving}

When FEAT is used without a complexity-aware survival method (i.e., with Lex, SimAnn, Random), a separate population is maintained that acts as an archive. The archive maintains a Pareto front according to minimum loss and complexity (Eqn~\ref{eq:complex}). At the end of optimization, the archive is tested on a small hold-out validation set. The individual with the lowest validation loss is the final selected model. Maintaining this archive helps protect against overfitting resulting from overly complex / high capacity representations, and also can be interpreted directly to help understand the process being modelled.

\begin{figure}
%%%%%%%%%%%%%%%%%%%%%%%%%%%%%%%%%%%%%%%%%%%%%%%%%%%%%%%%%%%%%%%%%%%%%%% Scores
\begin{minipage}{0.49\textwidth}    
    
    \centering
    \includegraphics[width=\textwidth]{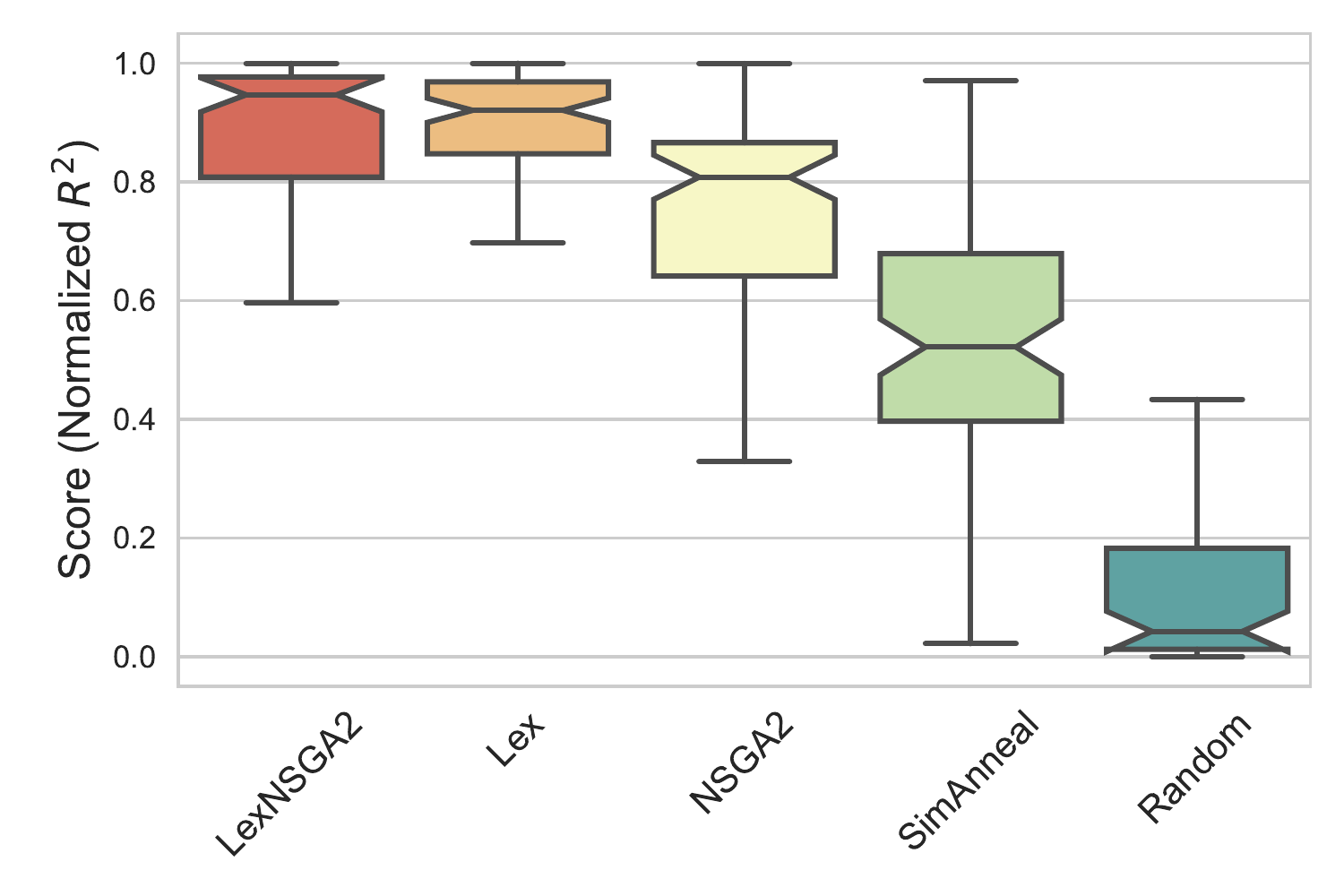}
    \caption{Mean 10-fold CV $R^2$ performance for various SO methods in comparison to other ML methods, across the benchmark problems.}\label{fig:score-so}

   \end{minipage}
\hspace{0.01\textwidth}
%%%%%%%%%%%%%%%%%%%%%%%%%%%%%%%%%%%%%%%%%%%%%%%%%%%%%%%%%%%%%%%%%%%%%%% Size
\begin{minipage}{0.49\textwidth}
    \centering    
    \includegraphics[width=\textwidth]{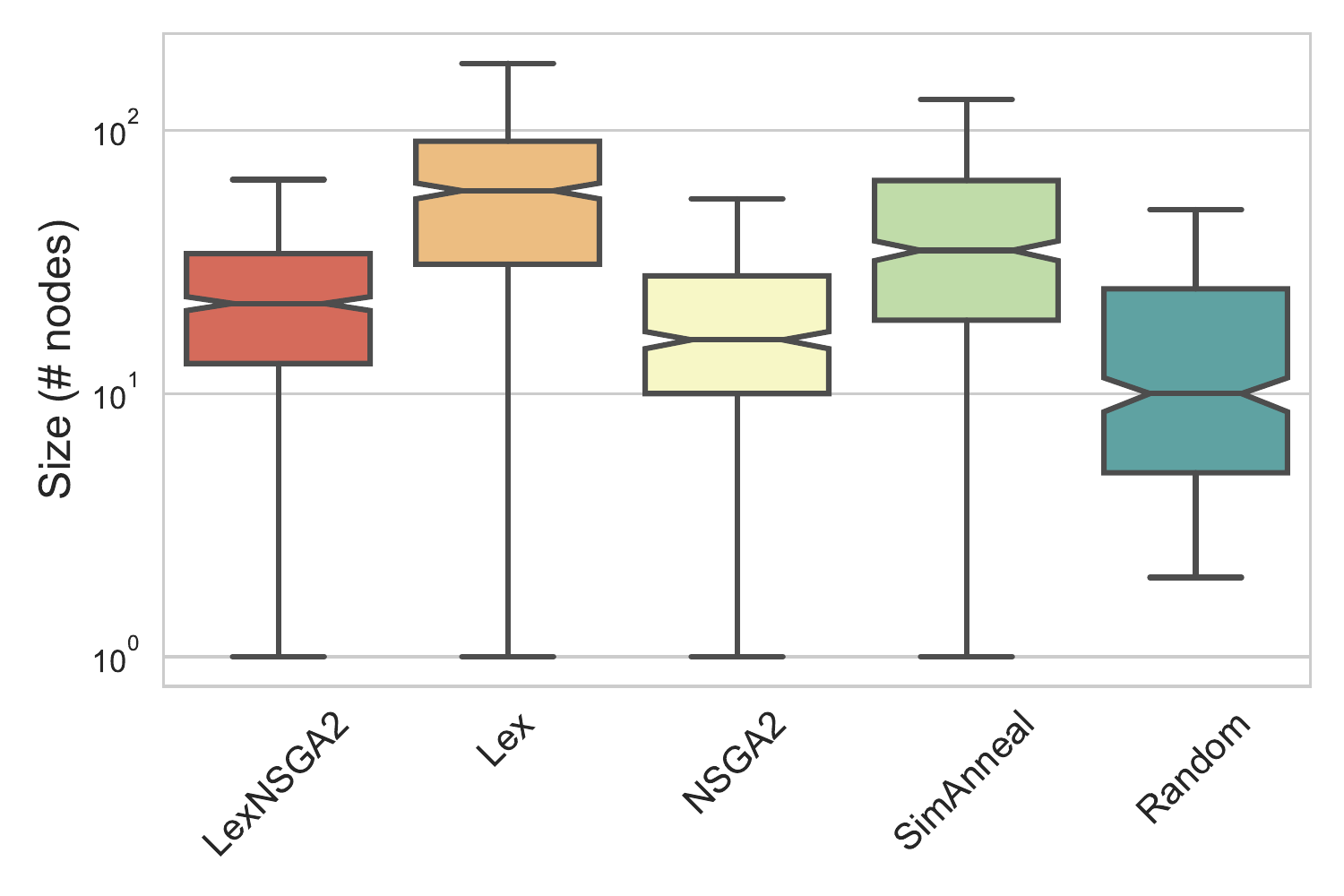}
    \caption{Size comparisons of the final models in terms of number of parameters. }\label{fig:size-so}
\end{minipage}
\end{figure}

\begin{figure}
\begin{minipage}{0.49\textwidth}
    \centering
    \includegraphics[width=\textwidth]{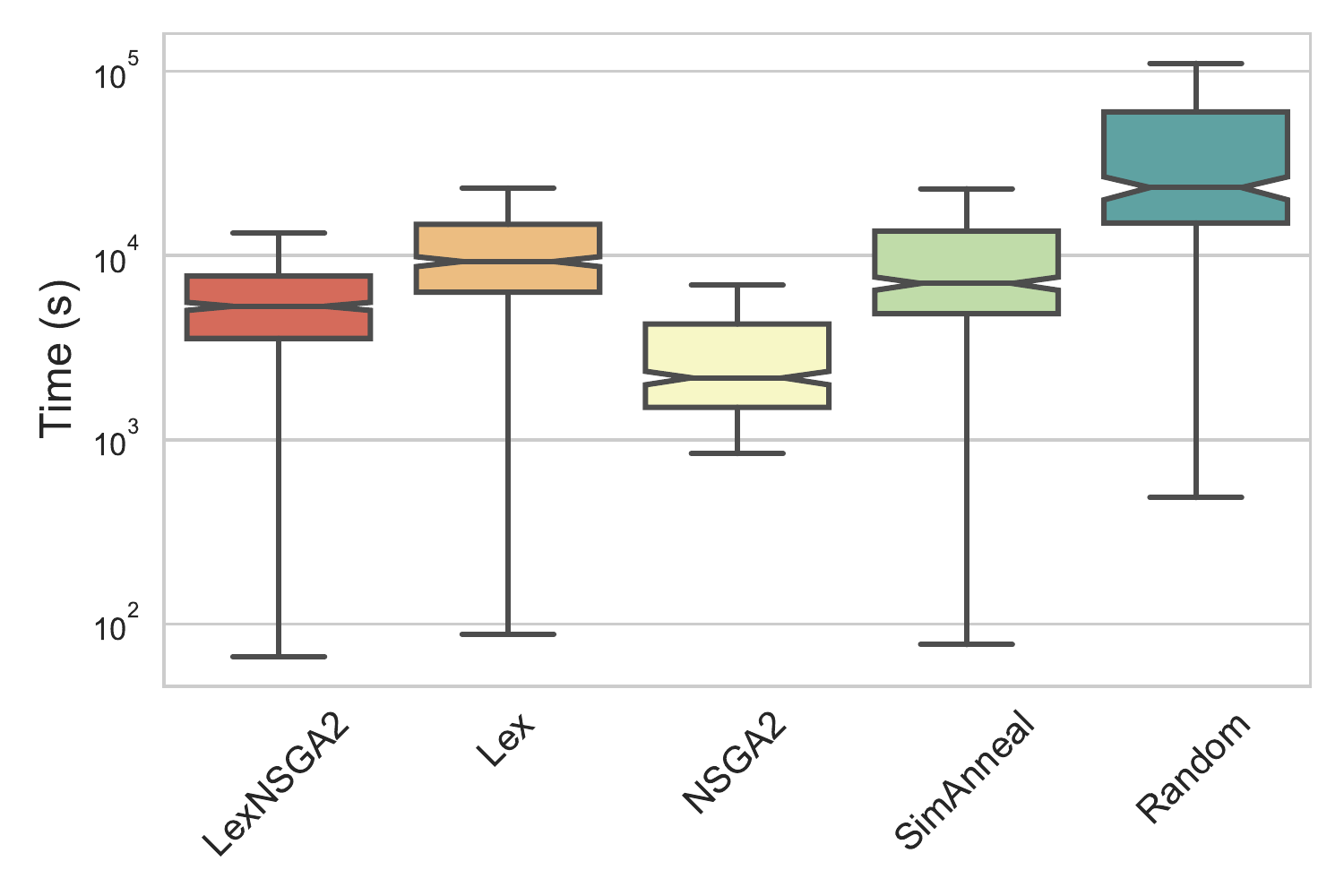}
    \caption{Wall-clock runtime for each method in seconds. }\label{fig:time-so}
\end{minipage}
\hspace{0.01\textwidth}
%%%%%%%%%%%%%%%%%%%%%%%%%%%%%%%%%%%%%%%%%%%%%%%%%%%%%%%%%%%%%%%%%%%%%%% Correlation
\begin{minipage}{0.49\textwidth}
    \centering
    \includegraphics[width=\textwidth]{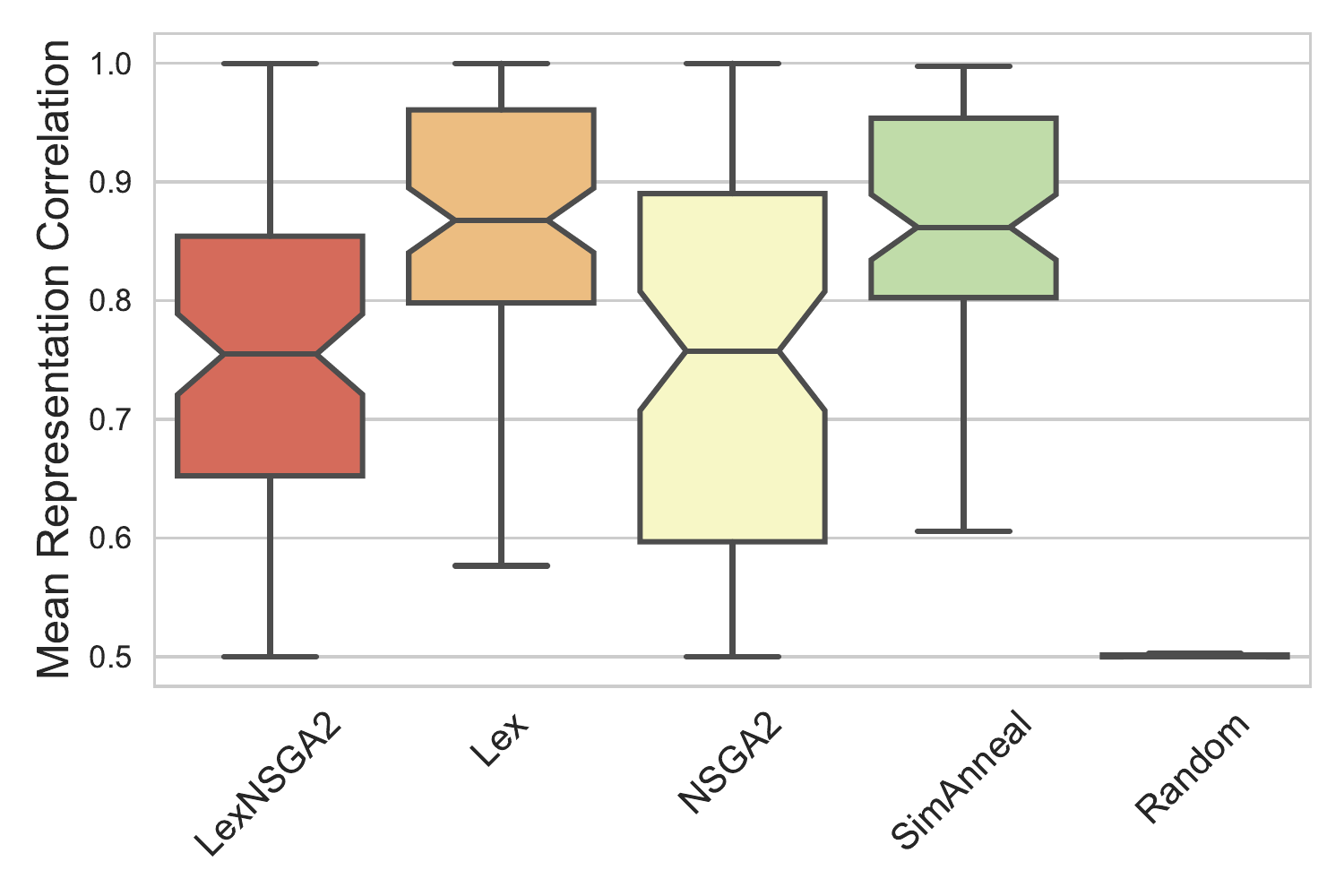}
    \caption{Mean correlation between engineered features for different SO methods compared to the correlations in the original data (ElasticNet). }\label{fig:corr-so}
\end{minipage}
\end{figure}

We benchmarked these approaches in a separate experiment on 88 datasets from PMLB~\citep{olson_pmlb:_2017}. The results are shown in Figures~\ref{fig:score-so}-\ref{fig:corr-so}. Considering Figures~\ref{fig:score-so} and~\ref{fig:size-so}, we see that LexNSGA2 achieves the best average $R^2$ value while producing small solutions in comparison to Lex. NSGA2, SimAnneal, and Random search all produce less accurate models. The runtime comparisons of the methods in Figure~\ref{fig:time-so} show that they are mostly within an order of magnitude, with NSGA2 being the fastest (due to its maintenance of small representations) and Random search being the slowest, suggesting that it maintains large representations during search. The computational behavior of Random search suggests the variation operators tend to increase the average size of solutions over many iterations. 

%%%%%%%%%%%%%%%%%%%%%%%%%%%%%%%%%%%%%%%%%%%%%%%%%%%%%%%%%%%%%%%%%%%%%%% Time

% \newpage
\subsection{Statistical Comparisons}\label{s:stats}
We perform pairwise comparisons of methods according to the procedure recommended by \cite{demsar_statistical_2006} for comparing multiple estimators. In Table~\ref{tbl:r2}, the CV $R^2$ rankings are compared. In Table~\ref{tbl:size}, the best model size rankings are compared. Note that KernelRidge is omitted from the size comparisons since we don't have a comparable way of measuring the model size.
% latex table generated in R 3.5.1 by xtable 1.8-3 package
% Fri Feb 22 00:41:16 2019
\begin{table}[ht]
\centering
\small
\caption{Bonferroni-adjusted $p$-values using a Wilcoxon signed rank test of $R^2$ scores for the methods across all benchmarks. *: $p$<0.05.}\label{tbl:r2}
\begin{tabular}{rlllllll}
  \hline
 & ElasticNet & Feat & FeatCN & FeatCorr & KernelRidge & MLP & RF \\ 
  \hline
Feat & 4.70e-14* &  &  &  &  &  &  \\ 
  FeatCN & 1.38e-12* & 5.34e-01 &  &  &  &  &  \\ 
  FeatCorr & 4.25e-13* & 1.00e+00 & 1.00e+00 &  &  &  &  \\ 
  KernelRidge & 1.16e-09* & 1.18e-04* & 4.37e-03* & 1.14e-03* &  &  &  \\ 
  MLP & 5.24e-09* & 3.80e-04* & 2.08e-02* & 1.28e-03* & 1.00e+00 &  &  \\ 
  RF & 1.08e-09* & 2.09e-07* & 2.19e-05* & 1.30e-06* & 1.00e+00 & 1.00e+00 &  \\ 
  XGB & 1.47e-13* & 1.00e+00 & 1.00e+00 & 1.00e+00 & 3.41e-04* & 1.60e-03* & 8.49e-13* \\ 
   \hline
\end{tabular}
\end{table}

% latex table generated in R 3.5.1 by xtable 1.8-3 package
% Fri Feb 22 00:42:09 2019
\begin{table}[ht]
\centering
\small
\caption{Bonferroni-adjusted $p$-values using a Wilcoxon signed rank test of sizes for the methods across all benchmarks. All results are significant. *: $p$<0.05.} \label{tbl:size}
\begin{tabular}{rllllll}
  \hline
 & ElasticNet & Feat & FeatCN & FeatCorr & MLP & RF \\ 
  \hline
Feat & 8.25e-12* &  &  &  &  &  \\ 
  FeatCN & 2.47e-16* & 2.12e-08* &  &  &  &  \\ 
  FeatCorr & 1.37e-12* & 1.00e+00 & 1.58e-07* &  &  &  \\ 
  MLP & 6.24e-18* & 4.26e-17* & 3.09e-17* & 3.98e-17* &  &  \\ 
  RF & 9.28e-20* & 2.05e-17* & 5.61e-18* & 2.37e-17* & 3.54e-17* &  \\ 
  XGB & 9.14e-18* & 4.05e-17* & 2.46e-17* & 3.71e-17* & 1.00e+00 & 3.94e-18* \\ 
   \hline
\end{tabular}
\end{table}

\end{document}